\documentclass[letterpaper,twocolumn,10pt]{article}
\usepackage{usenix20}
\usepackage[framemethod=tikz]{mdframed} 

\usepackage{xcolor}
\usepackage{xspace}
\usepackage{mathtools} 
\usepackage{amssymb}
\usepackage{bm}
\usepackage{enumitem}
\usepackage{textcomp}

\usepackage{graphicx}
\usepackage{subcaption}
\usepackage{tikz}
\usepackage[labelfont=bf]{caption}

\usepackage{tabularx}
\usepackage{multirow}
\usepackage{booktabs}
\usepackage{makecell}
\usepackage[figuresleft]{rotating}
\usepackage[flushleft]{threeparttable}
\usepackage{colortbl}

\usepackage{algorithmicx}
\usepackage{algorithm}
\usepackage[noend]{algpseudocode}

\algnewcommand\algorithmicinput{\textbf{Input:}}
\algnewcommand\Input{\item[\algorithmicinput]}
\algnewcommand\algorithmicoutput{\textbf{Output:}}
\algnewcommand\Output{\item[\algorithmicoutput]}

\usepackage[utf8]{inputenc} 
\usepackage{inconsolata} 
\usepackage[most]{tcolorbox} 
\usepackage{soul} 
\usepackage{siunitx} 
\usepackage{pifont} 
\usepackage{xurl} 
\usepackage{marginnote}
\usepackage{fontawesome5}

\usepackage[noindentafter]{titlesec}
\titlespacing\section{0pt}{3pt}{3pt} 
\titlespacing\subsection{0pt}{3pt}{3pt}
\titlespacing\subsubsection{0pt}{5pt}{5pt}

\usepackage[skip=3pt]{caption}
\newcommand{\TN}{\texttt{PolyTrace}\xspace}

\begin{document}

\date{}

\vspace{-50pt}
\newcommand{\papername}{RL in the Wild: Characterizing RLVR Training in LLM Deployment}

\title{\vspace{-10pt}\Large \bf \papername}


\author{
Jiecheng Zhou\textsuperscript{\ding{73}1}, Qinghao Hu\textsuperscript{2}, Yuyang Jin\textsuperscript{3}, Zerui Wang\textsuperscript{1,4}, Peng Sun\textsuperscript{5}, Yuzhe Gu\textsuperscript{1,4}
\\
Wenwei Zhang\textsuperscript{1}, Mingshu Zhai\textsuperscript{3}, Xingcheng Zhang\textsuperscript{1}, Weiming Zhang\textsuperscript{\ding{73}}
\vspace{5pt}
\\
{\it \textsuperscript{\ding{73}}USTC}
\qquad
{\it \textsuperscript{1}Shanghai AI Laboratory}
\qquad
{\it \textsuperscript{2}NTU}
\qquad
{\it \textsuperscript{3}Tsinghua University}
\\
{\it \textsuperscript{4}Shanghai Jiao Tong University}
\qquad
{\it \textsuperscript{5}Unaffiliated}
}

\maketitle

\begin{abstract}
  Large Language Models (LLMs) are now widely used across many domains. With their rapid development, Reinforcement Learning with Verifiable Rewards (RLVR) has surged in recent months to enhance their reasoning and understanding abilities. 
  However, its complex data flows, and diverse tasks pose substantial challenges to RL training systems, and there is limited understanding of RLVR from a system perspective. 
  To thoroughly understand the system challenges introduced by RLVR, we present a characterization study of RLVR tasks in our LLM deployment. 
  Specifically, we investigate the distribution and variation trends of workloads across different RL tasks across training steps. 
  We identify issues such as GPU idling caused by skewed sequence length distribution, inefficient parallel strategies in dynamically varying workloads, inefficient data management mechanisms, and load imbalance. 
  We describe our observations and call for further investigation into the remaining open challenges. 
  Furthermore, we propose \TN benchmark suite to conduct evaluation with realistic workloads, a practical use case validates that \TN benchmark suite exhibits 94.7\% accuracy.
\end{abstract}
\section{Introduction}
\label{sec_intro}


The rapid advancement of LLMs (e.g., DeepSeek\cite{deepseek}, OpenAI o1\cite{o1}, Gemini\cite{Gemini}) has fundamentally transformed the landscape of academia and industry. This exponential growth in model complexity has been accompanied by increasingly sophisticated training paradigms, evolving from traditional Supervised Fine Tuning (SFT) to more nuanced approaches that better improve model capabilities. The computational demands of training these massive models have necessitated the development of distributed training systems\cite{megatron, DeepSpeed,fsdp, Alpa} capable of handling huge scales of data and computation.

Within this broader context of LLM development, RLVR has emerged as a critical methodology for post-training optimization. Unlike conventional pretraining\cite{gpt3} and SFT\cite{SFT1,SFT2,instructGPT}, RLVR introduces a fundamentally different computational paradigm where models interact with environments, receive rewards, and iteratively refine their behavior through policy optimization algorithms such as Proximal Policy Optimization (PPO) \cite{PPO} and Group Relative Policy Optimization (GRPO) \cite{grpo}. This approach has proven essential for enhancing their reasoning capabilities\cite{rl_survey}, leading to widespread adoption in both research and industry. 

To enhance the performance of RLVR training systems, A thorough workload analysis is essential for comprehending challenges and uncovering opportunities. However, existing workload analysis conducted for other LLM tasks, such as pretraining\cite{inference_training_analyze, what_if} and serving\cite{servegen,modserve,kvcacheinthewild}, do not apply to RLVR training systems. This is primarily due to the divergent characteristics and requirements of RLVR:

\begin{table}[t]
\renewcommand{\arraystretch}{1.1}
\begin{tabular}{@{}cccc@{}}
\toprule
Workload       & Pretraining                & Serving                                   & RL                                               \\ \midrule
Prior Work     & \cite{MegaScale,bytescale, acme} & \cite{servegen,kvcacheinthewild,burstgpt} & -                                                \\
Step Time      & \faCar ($\sim$s)           & \faPlane ($\sim$ms)                       & \faWalking($\sim$min)                     \\
\#Model        & 1                          & 1                                         & 1$\sim$4                                         \\
\#Stage        & 2                          & 2                                         & 8+                                               \\
Static Data    & \ding{52}    & \ding{56}                   & \ding{56}                        \\
Heterogeneity   & \ding{56}    & \ding{56}                   & \ding{52}                         \\
Software Stack & \faStar[regular]           & \faStar[regular]                          & \faStar[regular]\faStar[regular]\faStar[regular] \\ \bottomrule
\end{tabular}
\caption{Comparison of Pretraining, Serving, and RL workload characteristics. RL workloads are the most complex, involving multiple stages, multiple models, dynamic data generation, and a tightly coupled software stack.}
\label{diff_rlvr}
\end{table}

\noindent \textbf{(1) \emph{Longer step span}}.
In RLVR, a single training step often takes on the order of minutes and, in some cases, can exceed one hour. By contrast, other LLM training tasks typically complete a step in several seconds, while an inference step (prefill and decode) is usually under 1s.

\noindent \textbf{(2) \emph{Complicated workflow}}.
In RLVR, a single training run requires multiple models to collaborate according to the algorithmic logic, and is often divided into several stages, such as rollout, inference, and training. By contrast, other LLM training tasks typically involve only one model and a single training stage\cite{instructGPT, Pretrain}. This introduces challenges to resource allocation and handling error propagation.

\noindent \textbf{(3) \emph{Unknown sequence length}}.
In prior LLM training tasks, the workload was known in advance, enabling offline reordering and scheduling policy selection\cite{bytescale}. In RLVR, however, the workload—much like that of LLM serving—cannot be predetermined and exhibits dynamics and abrupt shifts. This leads to suboptimal parallelization and difficulty in applying appropriate scheduling strategies.

\noindent \textbf{(4) \emph{Complex software stack}}.
Previously, LLM tasks could be executed within a single training or inference framework, such as Megatron\cite{megatron}, FSDP\cite{fsdp}, vLLM\cite{vllm}, or SGLang\cite{sglang}. In RLVR, however, the training framework must be built atop existing training and inference frameworks, and additionally incorporate software for environment interaction, such as sandboxes\cite{sandbox_fusion} and MCP-based\cite{mcp} programs. 

To bridge this gap, we present \TN and an in-depth analysis of RLVR training systems based on our operational experiences across multiple production tasks. To the best of our knowledge, this represents the first comprehensive study of RLVR workloads in real-world deployment scenarios. Our analysis draws upon extensive traces collected from large-scale RLVR training jobs, encompassing detailed profiling data and system performance logs. Our further key findings can be summarized as follows:

\begin{itemize}[leftmargin=*,topsep=0pt, itemsep=-3pt, itemindent=8pt]
\item \textrm{\bfseries Various and long-tail sequence length distributions.} Different RL tasks exhibit substantial variations in input and output sequence length distributions. In terms of output length, mathematics and image understanding tasks generate extremely long output sequences, with average lengths reaching 9,839.3 and 5,911.6 tokens, respectively, whereas video understanding and searching tasks exhibit average output lengths of only 166.5 and 48.7 tokens. The pattern for input lengths is markedly different: due to the presence of video information, video understanding tasks have an average input length of 3,123.1 tokens, while mathematics tasks require merely 126.5 tokens. Meanwhile, long-tail distributions are prevalent across RL tasks. For instance, in tool-use tasks, the 90th percentile versus the maximum input length differs by up to 18×, while the 90th percentile versus the maximum output length reaches approximately 5×.

\item \textrm{\bfseries Dynamically varying output length during training.} On one hand, the variation trends in output length are model-dependent: larger models typically do not exhibit pronounced output length changes, whereas smaller models demonstrate significant sequence length variation trends. On the other hand, different tasks also show markedly distinct variation patterns. For example, searching tasks exhibit decreasing output lengths as training progresses, while mathematics tasks show increasingly longer outputs.

\item \textrm{\bfseries Fluctuating system performance during training process.} In large-scale RL training tasks, performance varies dramatically across training steps. For instance, in image understanding tasks, the best performance can reach 341 tokens per GPU per second (TGS), while the worst performance drops to merely 0.8 TGS. This instability stems from diverse sequence lengths and inappropriate parallelization strategies.
\end{itemize}

Based on our characterization study, we identify several challenges and insights in RL training systems. \emph{Firstly}, workload-aware request scheduling strategies are essential. The diverse sequence length distribution characteristics significantly impact RL training system performance, and simple scheduling strategies such as First Come First Serve (FCFS) cannot achieve satisfactory performance in RL training. SLO-aware scheduling strategies(e.g., prefill and decoding disaggregation\cite{DistServe}) are not profitable in RL training. \emph{secondly}, inefficient data transmission and management implementations. To improve framework usability, RL training frameworks often adopt single-controller designs, but in certain tasks, this design creates single-point bottlenecks that lead to degraded training performance or even failure. \emph{thirdly}, severe load imbalance. The unbalanced sequence length distributions in RL training result in uneven, uncertain, and dynamic sequence length during the training stage; current static parallelization strategies and coarse-grained load balancing algorithms exhibit poor performance. Other challenges, such as expensive scheduling overhead in inference frameworks, coarse-grained and inefficient GPU memory management, and uncertain tool invocation latencies, indicate that RL training systems require further design and optimization. The benchmark suite serves as a crucial tool for uncovering system insights in new cases and enabling researchers to validate their optimization methods effectively. To this end, we propose \TN and \TN benchmark suite to uncover system insights and comprehensive evaluations with realistic workloads. A practical use case validates that \TN benchmark suite exhibits 94.7\% accuracy.

We believe the observations and insights from our work are broadly applicable. 
Based on our analysis, we highlight the profound impact of workload distribution on the performance of RL training systems. 
Consequently, our traces and benchmark suite are publicly available at \burl{https://github.com/Zhou-jiecheng/PolyTrace} to facilitate further research and development in this area.
\section{Background}
\label{sec_background}

\subsection{LLM Reinforcement Learning}
\begin{figure}[t]
    \centering
    \includegraphics[width=\linewidth]{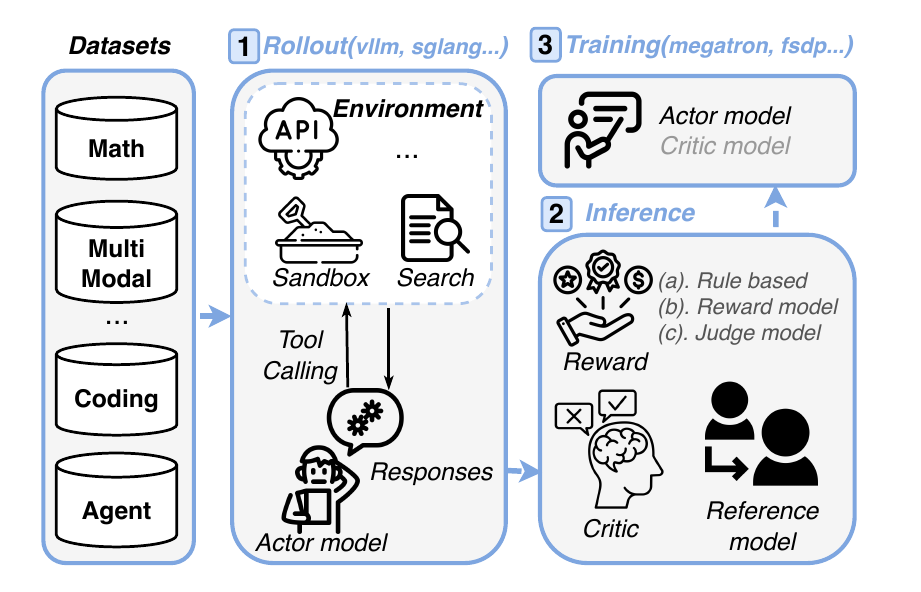}
    \caption{Overview of the RL Training Workflow.}
    \label{rl_workflow}
\end{figure}

\textbf{RL Training Workflow.}
As illustrated in Figure \ref{rl_workflow}, unlike pre-training and SFT, RLVR typically comprises three stages and multiple LLMs. The workflow can be divided as follows:

\textbf{\emph{1. Rollout.}} The actor model, analogous to an agent model in the RL setting, performs autoregressive generation\cite{Attention17} starting from an initial state. Each generated token constitutes an action that transitions to a new state. Once generation terminates, the resulting sequence is treated as a trajectory for subsequent stages. This stage is usually executed by inference engines (e.g., vLLM\cite{vllm}, SGLang\cite{sglang}, and LMdeploy\cite{turbomind}). The actor model may also invoke external tools (e.g., image processing\cite{deepeyes}) in rollout.

\textbf{\emph{2. Inference.}} This stage may involve multiple models. At a minimum, it includes forward passes of the actor model and a reference model, typically orchestrated by the training framework\cite{megatron,fsdp, DeepSpeed}. To obtain the reward for the actor model, different RL tasks have diverse choices, commonly falling into three categories: (a) Rule-based rewards. Directly match the response produced by the actor model against predefined rules to assign a reward score. This is suitable for tasks with readily verifiable correctness, such as coding or mathematics. (b) Reward model-based rewards: When the reward cannot be formalized, a trained reward model—often a fine-tuned base LLM—provides the reward score via a forward pass. (c) Judge-model-based rewards: To avoid the substantial cost of training a reward model, some methods directly use a stronger LLM (often a larger one) to evaluate the response and produce a reward score. Unlike the previous cases, this step is commonly executed via an inference engine. The critic in the GRPO algorithm is a relative value calculated based on a set of responses from the same prompt, whereas in the PPO algorithm, it is provided through a critic model.

\textbf{\emph{3. Training.}} The actor model and, in PPO, the critic model are trained to update model parameters. This stage resembles pretraining and supervised fine-tuning (SFT) in that it is handled directly by LLM training frameworks such as Megatron\cite{megatron} or FSDP\cite{fsdp}. Training LLMs efficiently at scale necessitates various system innovations, such as state sharding optimizers\cite{Zero}, meticulous model placement using data\cite{ddp}, pipeline\cite{PipeDream}, and tensor parallelisms.

Because training objectives differ across tasks \cite{rl_survey}, models "react" differently to each task, resulting in highly diverse workloads. In such a multi-stage pipeline, workload characteristics exert a pronounced impact on the system.

\noindent \textbf{RL training system}
The RL training systems can be primarily categorized into synchronous and asynchronous. The core distinction lies in whether there exists staleness in the model version in the rollout stage of the RL training process.

\begin{figure}[t]
    \centering
    \includegraphics[width=\linewidth]{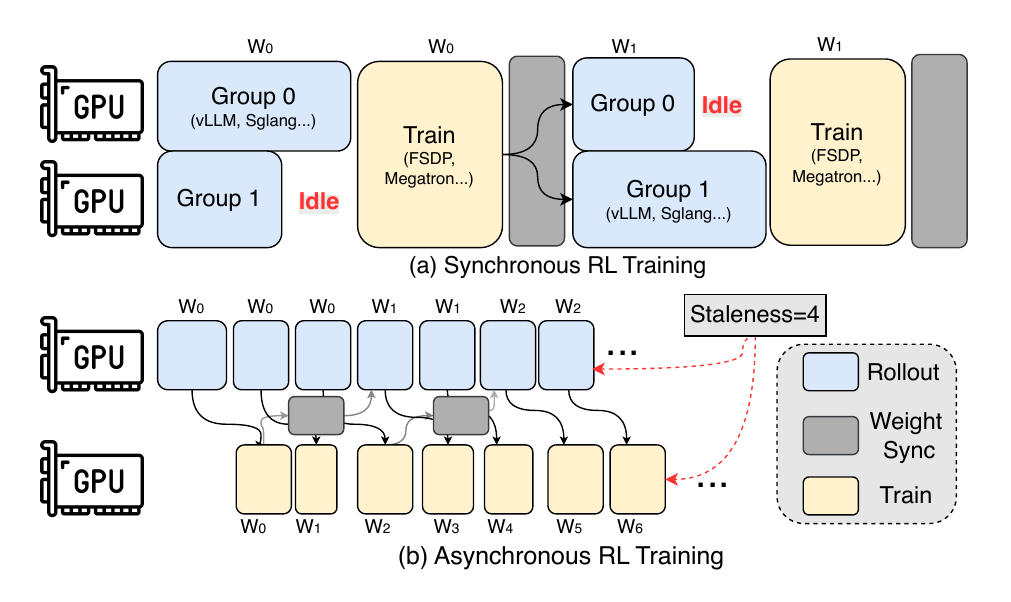}
    \caption{Comparison of Synchronous and Asynchronous RL Training. We omit additional stages (e.g., inference) to better highlight the fundamental distinction between synchronous and asynchronous RL training.}
    \label{rl_async}
\end{figure}

\textbf{\emph{1. Synchronous RL training.}} As illustrated in Figure \ref{rl_async} (a), in Synchronous RL training, each phase strictly respects data dependencies; consequently, rollout and training are executed with identical parameter versions. While straightforward, this strategy incurs severe GPU under-utilization when the workload is highly skewed, limiting its scalability.

\textbf{\emph{2. Asynchronous RL training:}}
Recent algorithmic studies\cite{areal} relax these dependencies by allowing the rollout phase to lag behind training by up to a \emph{maximum permitted staleness}. This pipelining reduces GPU idle time (Figure \ref{rl_async} (b)). Moreover, the two phases exhibit distinct compute characteristics: rollout is usually memory-bound due to long generation lengths, whereas training is compute-bound. Exploiting this disparity enables the use of heterogeneous hardware\cite{streamrl}, improving training efficiency and lowering cost.

\subsection{\TN Overview}
\label{sec_failure}

Motivated by prior discussions and the evident gap in publicly available traces, we introduce \TN to fill this void. As shown in Table \ref{tab:rl-traces}, \TN comprises workloads from seven representative reinforcement learning (RL) tasks, including four open-source RL training tasks and three large-scale RL training tasks used in model development. Each item entry captures the following fields:

\begin{enumerate}[itemsep=0pt, topsep=2pt, parsep=2pt, partopsep=2pt]
  \item \textbf{Step.} The global training step to which the record belongs.
  \item \textbf{Input length.} To preserve anonymity for certain internal datasets, we disclose only the input length for each input.
  \item \textbf{Output length.} The length of the model output associated with each prompt.
  \item \textbf{Type.} The task category of the workload record.
\end{enumerate}

Unlike serving traces \cite{mooncake, burstgpt}, RL training workloads intentionally omit arrival timestamps. 
In RL training, samples within the same step can be treated as arriving concurrently. 
\TN serves two complementary purposes: on one hand, our comprehensive trace is designed to catalyze systems research on RL training optimization, on the other hand, it facilitates rigorous, apples-to-apples evaluation of RL frameworks under diverse tasks.
\section{Characterize RL Training in the Wild}
\label{sec_characterization}
In this section, we present a thorough analysis of RL training workloads collected from RL Training logs 
during model production. Our study includes three kinds of RL tasks: image understanding for model A 
in Section \ref{mm_understand_workload}, mathematics and tool use (Section \ref{tooluse_workload}) for 
model B (Section \ref{math_task_workloads}). The training configurations are detailed in Table \ref{train_config}

\subsection{Vision-Task Workloads}
\label{mm_understand_workload}

\begin{figure}[t]
    \centering
    \includegraphics[width=\linewidth]{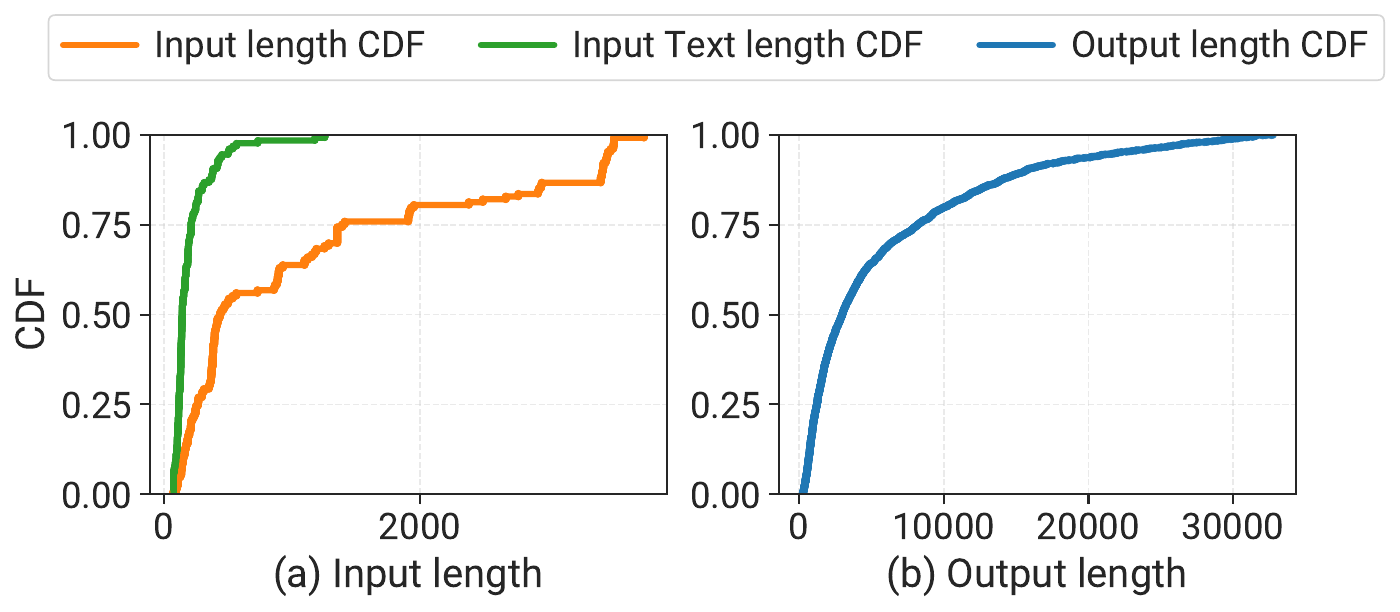}
    \caption{CDF of \emph{Image Understanding Task} input and output length distribution}
    \label{mm_input_output_cdf}
\end{figure}

\begin{figure}[t]
    \centering
    \includegraphics[width=\linewidth]{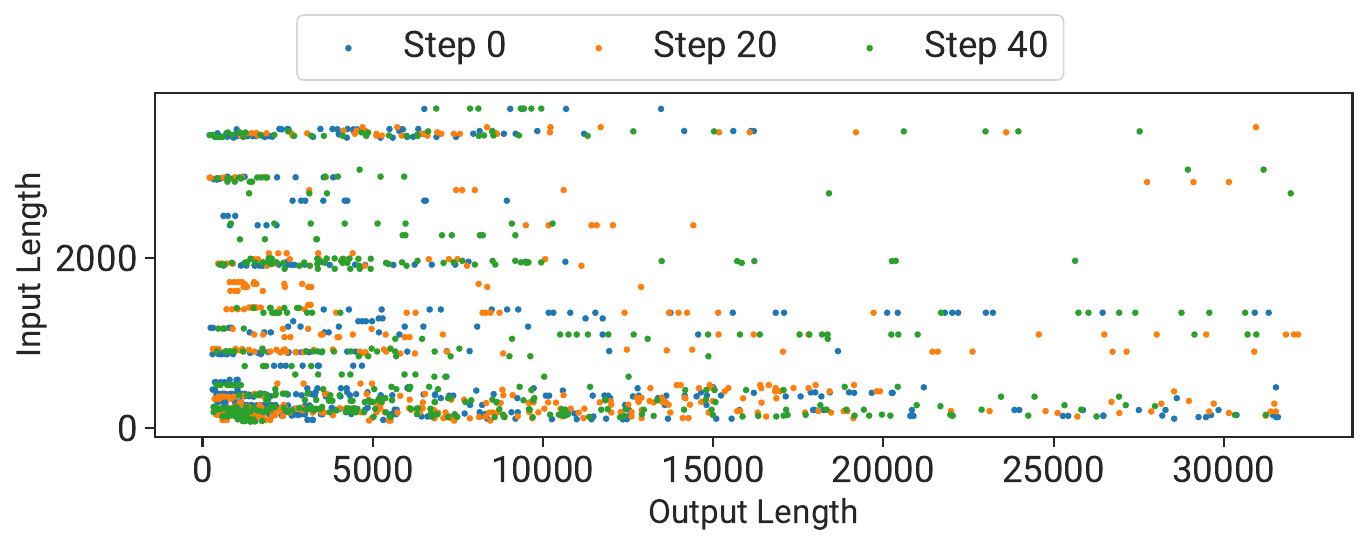}
    \caption{Scatter of \emph{Image Understanding Task} input and output length joint distribution}
    \label{mm_input_output_scatter}
\end{figure}

\begin{figure}[t]
    \centering
    \includegraphics[width=\linewidth]{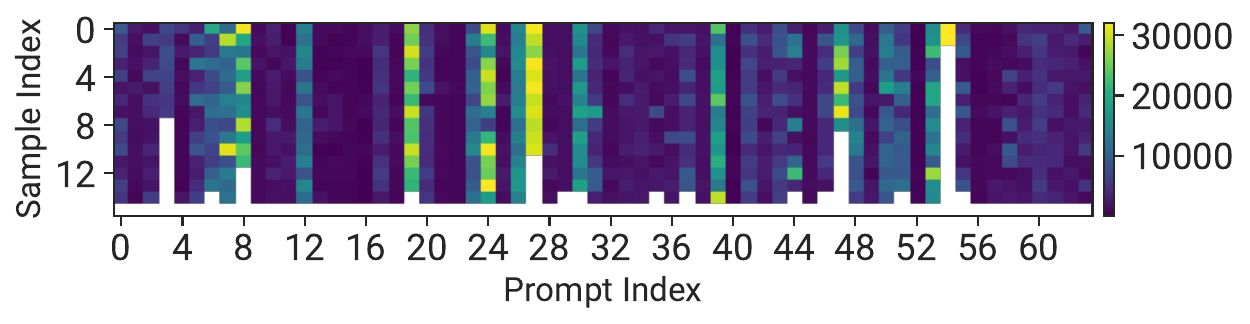}
    \caption{Heatmap of \emph{Image Understanding Task} prompt and outputs, we randomly sample 64 prompts for readability}
    \label{mm_prompt_heatmap}
\end{figure}

\begin{figure}[t]
    \centering
    \includegraphics[width=\linewidth]{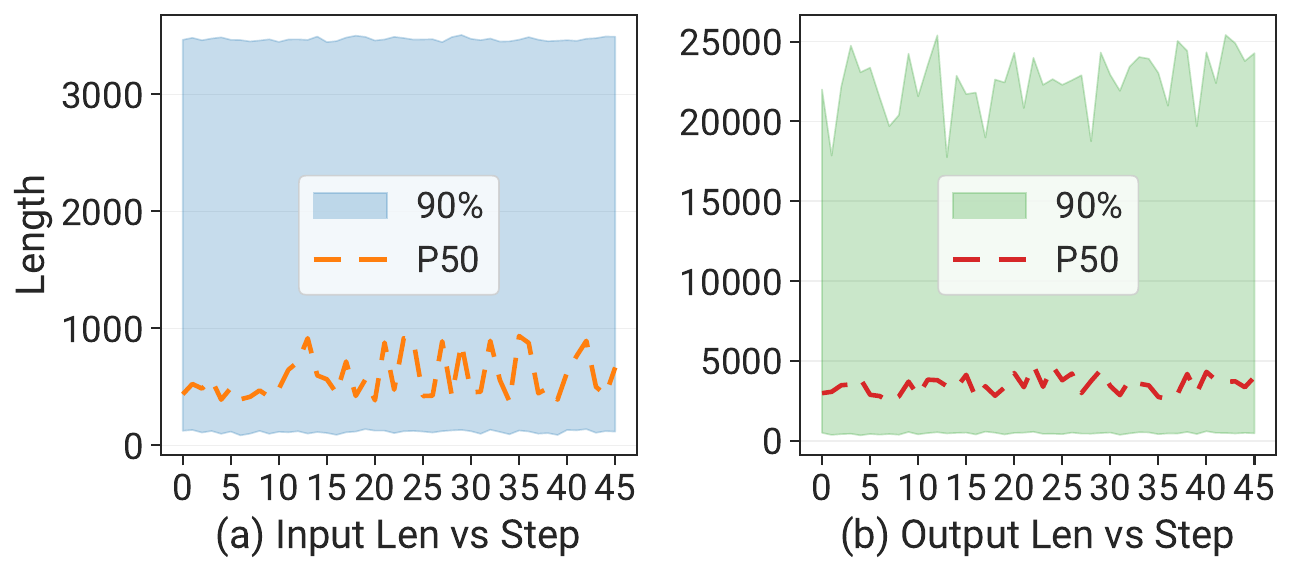}
    \caption{The workloads distribution across training steps of \emph{Image Understanding Task}.}
    \label{mm_step_distribution}
\end{figure}

\begin{figure}[t]
    \centering
    \includegraphics[width=\linewidth]{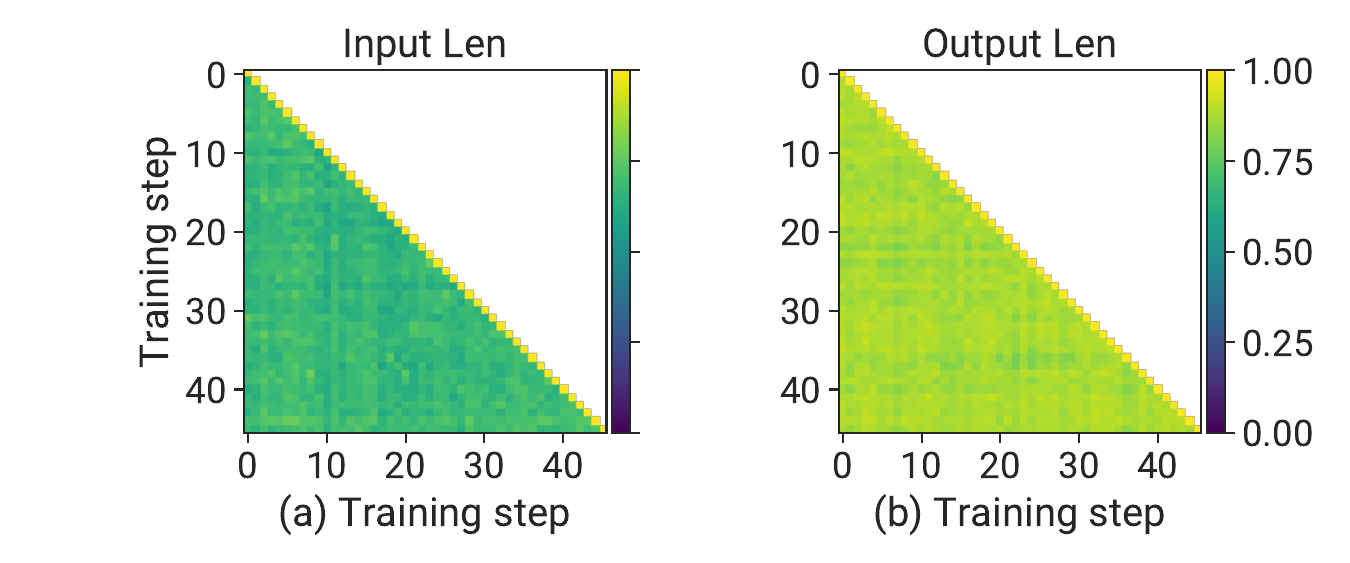}
    \caption{Similarity of distribution across training steps of \emph{Image Understanding Task}. The metric in the figure is 1 minus the Jensen–Shannon (JS) divergence.}
    \label{mm_similarity}
\end{figure}

\begin{figure}[t]
    \centering
    \includegraphics[width=\linewidth]{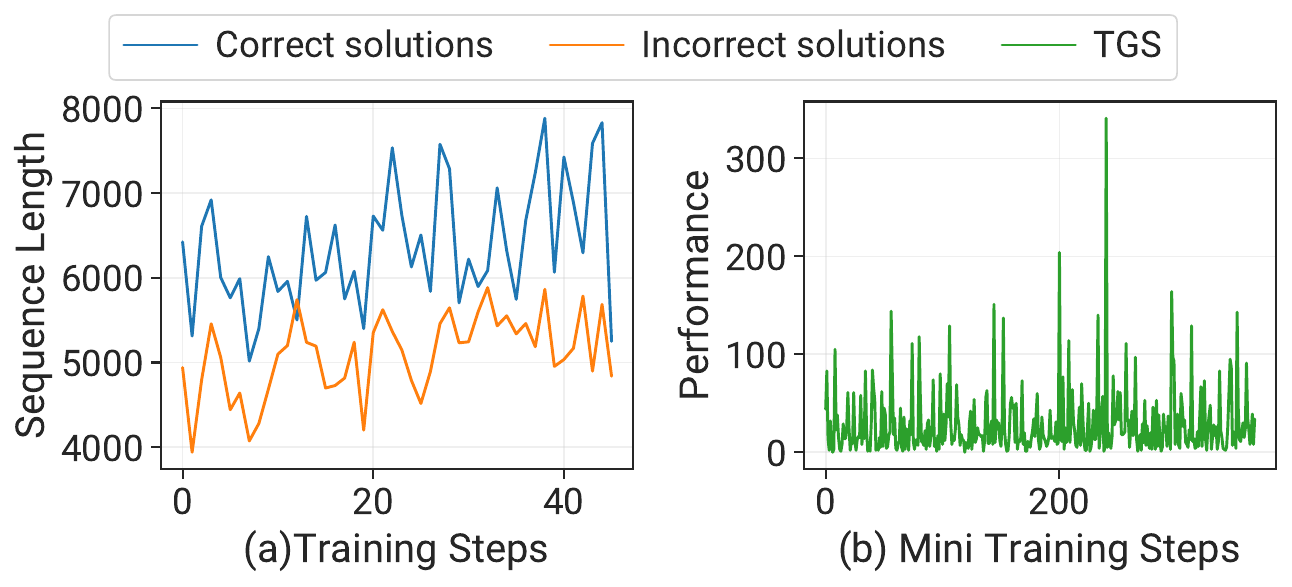}
    \caption{Figure (a) shows the average sequence lengths of correct and incorrect answers. Figure (b) depicts the performance at each mini training step during the training stage, where the performance metric is Tokens per GPU per Second (TGS). In RL training, once the rollout completes the generation of data for a global batch, the algorithm partitions this global batch into several mini-batches (8 in this case).}
    \label{mm_tgs}
\end{figure}

\begin{table}[t]
\centering
\renewcommand{\arraystretch}{1.25}
\resizebox{\linewidth}{!}
{
\begin{tabular}{@{} l l l r l l l l @{}}
\toprule
Model & Arch & Param & GPUs & Placement & Train Par & Rollout Par \\
\midrule
A & MoE & 235B & 256 & Colocate &
\begin{tabular}[t]{@{}c@{}}FSDP, SP=4\end{tabular} &
EP=16 \\
\midrule
B & MoE & 235B & 128 & Colocate &
\begin{tabular}[t]{@{}c@{}}TP=4, ETP=1\\EP=8, PP=8\end{tabular} &
TP=8 \\
\bottomrule
\end{tabular}
}
\caption{Training Configuration. Both of them use FP8 parameters during the rollout stage. We did not scale the training to a very large size due to the poor scalability of RL training (Section \ref{scale_study}).}
\label{train_config}
\end{table}

\textbf{Input/Output length characteristics.} As shown in Figure \ref{mm_input_output_cdf}, we first investigate the input and output length distributions for the multimodal content understanding task. The inputs exhibit a stepwise pattern driven by the varying number of images per sample, while the output length display a moderately shifted long tail: the median and 90th percentile are 2{,}978.5 and 15{,}601.8 tokens, respectively, with a maximum of 32{,}000 tokens, reflecting reasoning-oriented prompt design that encourage extended responses during RL training. These differ markedly from prior observations in LLM serving traces\cite{servegen, modserve}, where workloads tend to feature long inputs and short outputs. Meanwhile, the ratio between multimodal and text input lengths spans from 0 to 39.11. Overall, multimodal content accounts for the majority of the input length (mean ratio 8.57$\times$). This is because our carefully curated dataset covers multiple fields of knowledge, and some areas like geography include substantial amounts of complex visual information. This highlights the urgent need for decode phase optimization, which usually is the most time-consuming part (Section \ref{time_breakdown}).

\textbf{Joint input–output structure.}
We further examine the input–output joint distribution. As shown in Figure \ref{mm_input_output_scatter}, the distribution exhibits a banded pattern, corresponding to the input-distribution pattern in Figure \ref{mm_input_output_cdf}. There is no pronounced correlation between input and output lengths, only around an input length of roughly 2k do long-tail outputs become less likely. The potential factor is that input length is largely decoupled from prompt difficulty and will not affect the model’s reasoning length. Across steps, the joint distributions remain strongly similar, reflecting the difficulty distribution of prompts sampled from the dataset is similar across steps.

\textbf{Prompt-correlated output lengths.}
In RL training, algorithms typically perform multiple stochastic samples (rollouts) for the same prompt, producing a set of outputs. As illustrated in Figure \ref{mm_prompt_heatmap}, each prompt is sampled 16 times; darker colors denote shorter outputs, while blank cells indicate that the algorithm filtered out low-quality training samples. We observe strong similarity in output lengths across samples for the same prompt. This implies that per-sample output length is not fully random: for the same prompt, the model tends to produce outputs of comparable length, likely because the token budget required to solve the same underlying problem is similar across samples. As a result, we recommend scheduling at the prompt granularity rather than per sample. Also, it is possible to first run a preliminary sampling pass to obtain sample output lengths, then schedule other samples accordingly, rather than relying on an additional model to predict the output length\cite{streamrl}.

\textbf{Similar temporal distribution.}
In Figure \ref{mm_step_distribution}, we further examine how the input and output length distributions evolve across training steps. As training progresses, the medians of input and output lengths exhibit only minor jitter without clear temporal trends. However, output-length distributions can fluctuate more substantially across steps: for example, at step 13, the 95th percentile is around 16k, whereas at steps 41 and 12 it reaches roughly 25k. Moreover, Figure \ref{mm_similarity} shows that output-length distributions across steps are strongly similar, while input distributions lack clear cross-step similarity. This is primarily due to the carefully designed prompt sampling strategy and the diversity of the datasets. Last but not least, RL often operates with very few epochs—indeed, for Model~A, the epoch count can be less than one—rendering epoch-based optimizations (e.g., cross-epoch data reuse) less effective and of limited generality. In Figure \ref{mm_tgs} (a), we can observe that samples with correct answers generally have longer sequence lengths than those with incorrect answers, across training steps. which may indicate tha,t as training progresses, more correct responses will lead to increased sequence lengths.

\textbf{Training performance volatility.}
As illustrated in Figure \ref{mm_tgs} (b), during the training stage, performance exhibits significant fluctuations, with the maximum TGS reaching 341 while the minimum TGS is only 0.8 (~400x difference). Based on our in-depth analysis, there are two primary contributing factors: (1) The input-output length distribution (Figure \ref{mm_input_output_cdf}) and sample filtering strategy\cite{oreal} result in substantial variations in sequence lengths across mini training steps. For instance, at the step where TGS is 0.8, the training samples have a sequence length of only 435, whereas at the step where TGS reaches 341, the training samples have a sequence length of 31,280. Longer sequence lengths entail greater computational workload, thereby overlapping the overhead introduced by communication. (2) Static parallelization strategies fail to achieve satisfactory performance. Throughout this training process, we consistently employed the same parallelization configuration with \texttt{sequence parallel = 4}, which severely impacts training throughput. When sequence lengths are short, using sequence parallel\cite{jacobs2023deepspeed} introduces additional communication overhead instead. This observation underscores the significance of an online sequence-length-aware parallelization strategy in RL training.

\subsection{Mathematics-Task Workloads}
\label{math_task_workloads}

\begin{figure}[t]
    \centering
    \includegraphics[width=\linewidth]{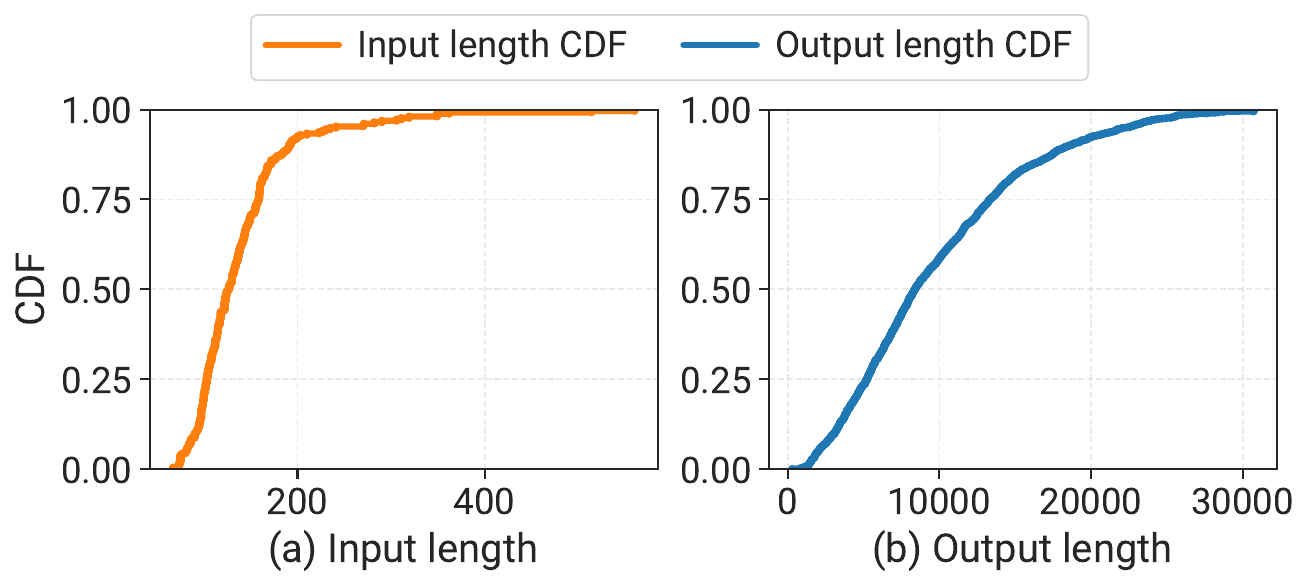}
    \caption{CDF of \emph{Mathematics Task} input and output length distribution}
    \label{math_input_output_cdf}
\end{figure}

\begin{figure}[t]
    \centering
    \includegraphics[width=\linewidth]{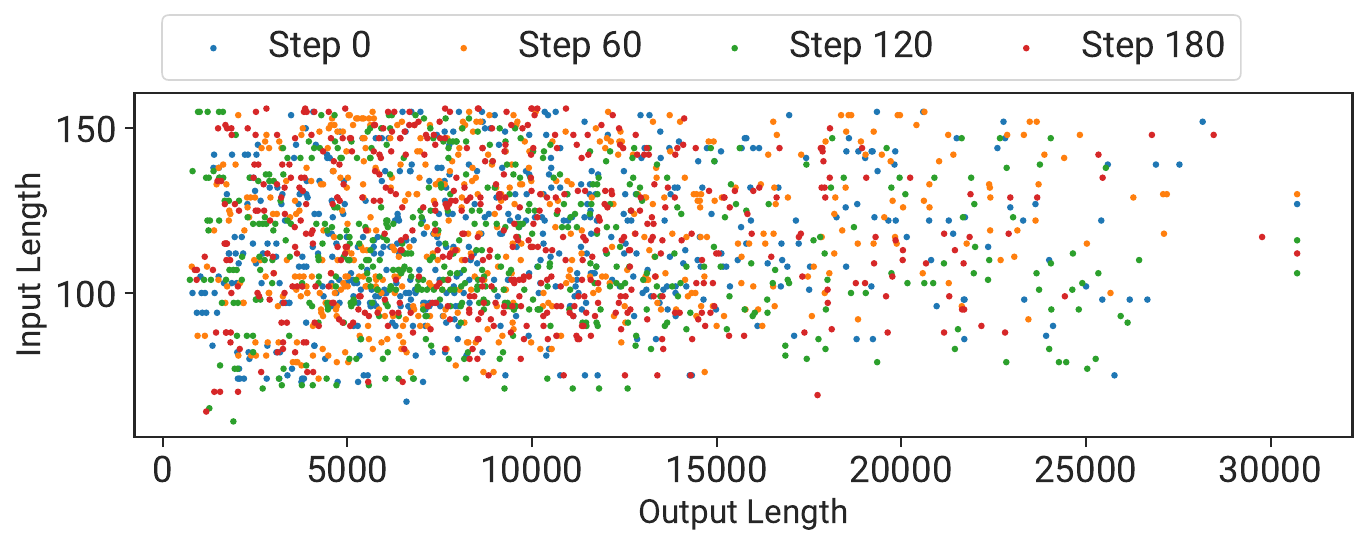}
    \caption{Scatter of \emph{Mathematics Task} input and output length joint distribution}
    \label{math_scatter}
\end{figure}

\begin{figure}[t]
    \centering
    \includegraphics[width=\linewidth]{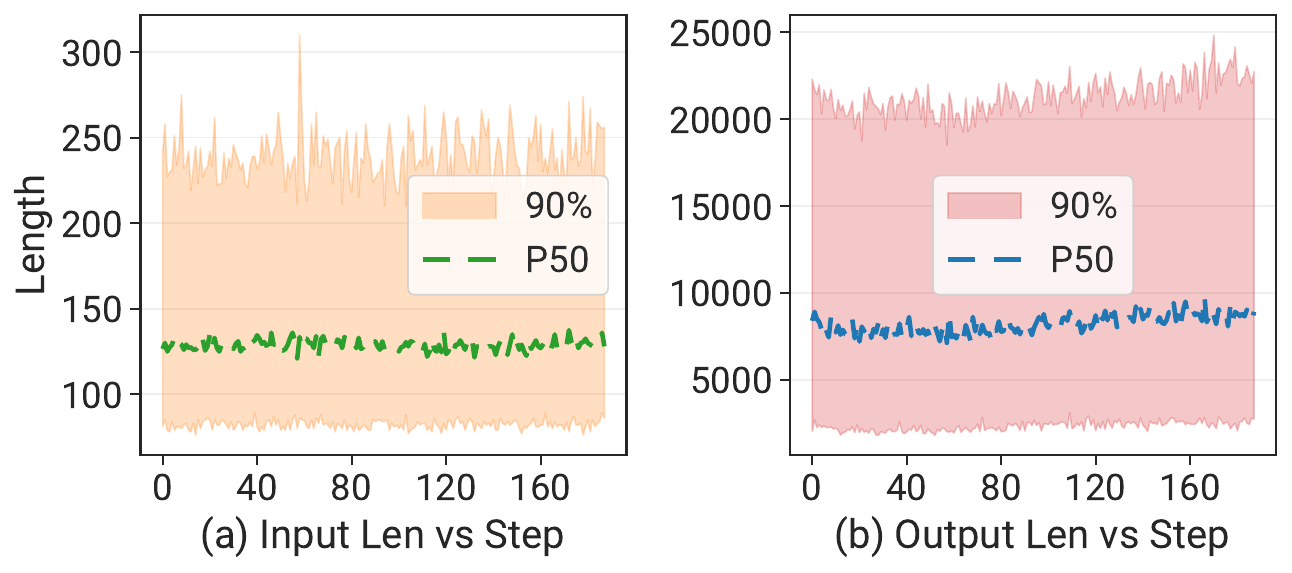}
    \caption{The workloads distribution across training steps of \emph{Mathematics Task}.}
    \label{math_step_distribution}
\end{figure}

To strive for a deeper understanding of the characteristics of different workloads in RL training during model production, we further conduct comprehensive analysis of Model B.

\textbf{Input/output length distribution characteristics.}
We further analyze the input–output length characteristics for math tasks. As shown in the Figure \ref{math_input_output_cdf}, compared with multimodal content understanding, math exhibits a smoother long-tailed output-length distribution: the 90th percentile reaches 18{,}543.1 tokens and the median reaches 8{,}388.0 tokens, while the longest sample is 32k tokens. Overall outputs are substantially longer—the median is 2.8$\times$ that of the multimodal task—primarily because RL-trained math models need more deliberate reasoning to solve challenging problems. In contrast, 90\% of math-task inputs are shorter than 200 tokens, and the longest input length is only about 600 tokens, which differs a lot from \emph{image understanding task}. This discrepancy arises because math inputs are typically standalone problems. Moreover, as illustrated in Figure \ref{math_scatter}, the joint input–output distribution forms an apparently patternless scatter, which indicates that input length is not correlated with problem difficulty, similar to \emph{image understanding task}, predicting output length must incorporate auxiliary information like prompt semantics.

\textbf{Stability of distributions over training.}
We next examine how these distributions evolve during training in Figure \ref{math_step_distribution}. The input and output length distributions for math show no pronounced temporal trend; however, we observe a very gradual increase in output length. This likely reflects improving reasoning ability as RL progresses, which elongates the chain of thought. Due to the very large parameter scale and the strong capabilities imparted by pretraining and SFT, large models under RL do not exhibit the pronounced lengthening seen in smaller models as displayed in Section \ref{workload_distribution}.

\subsection{Tool-use Task Workloads}
\label{tooluse_workload}
\begin{figure}[t]
    \centering
    \includegraphics[width=\linewidth]{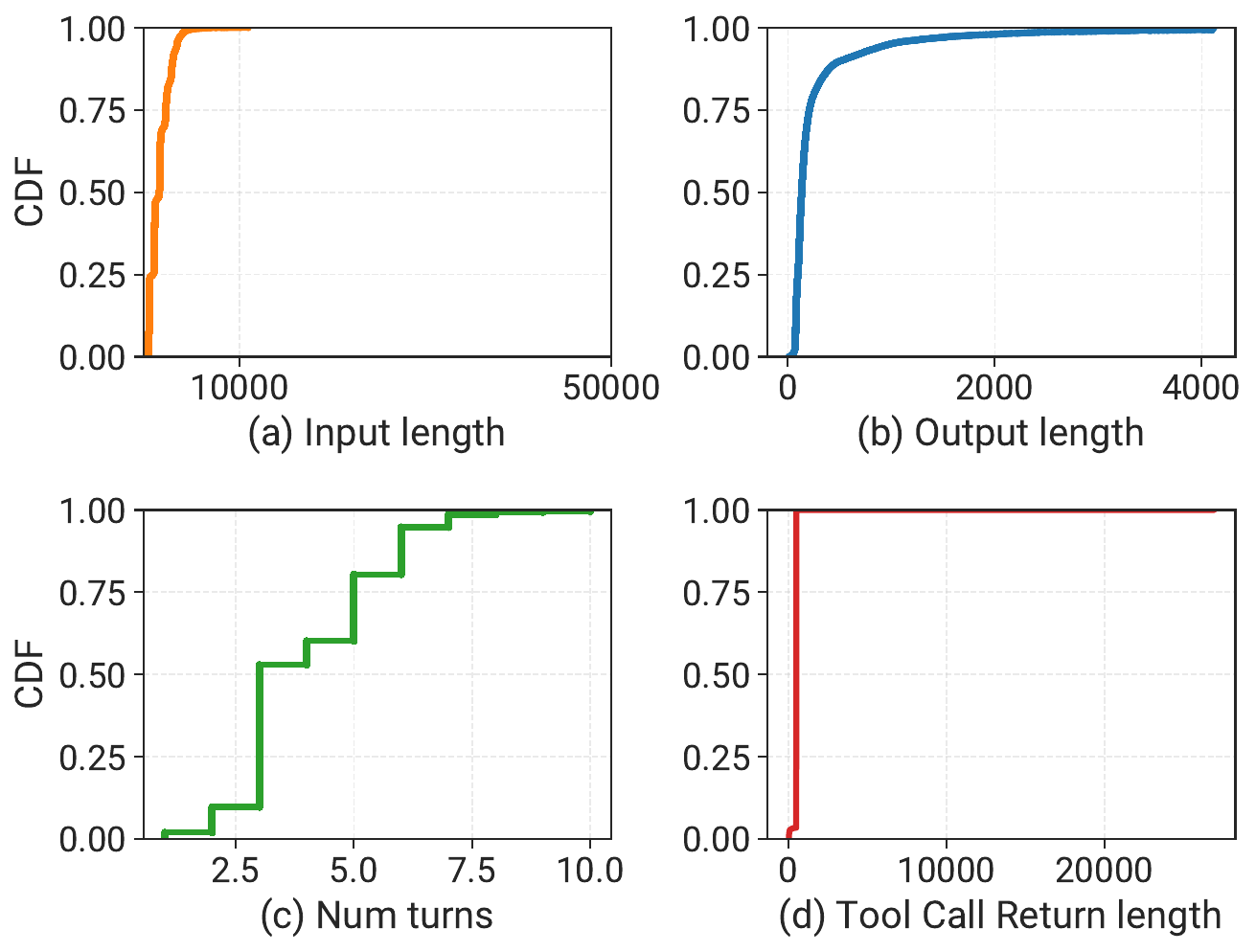}
    \caption{CDF of \emph{Tool-use Task} input and output length distribution}
    \label{tool_use_input_output_cdf}
\end{figure}

\begin{figure}[t]
    \centering
    \includegraphics[width=\linewidth]{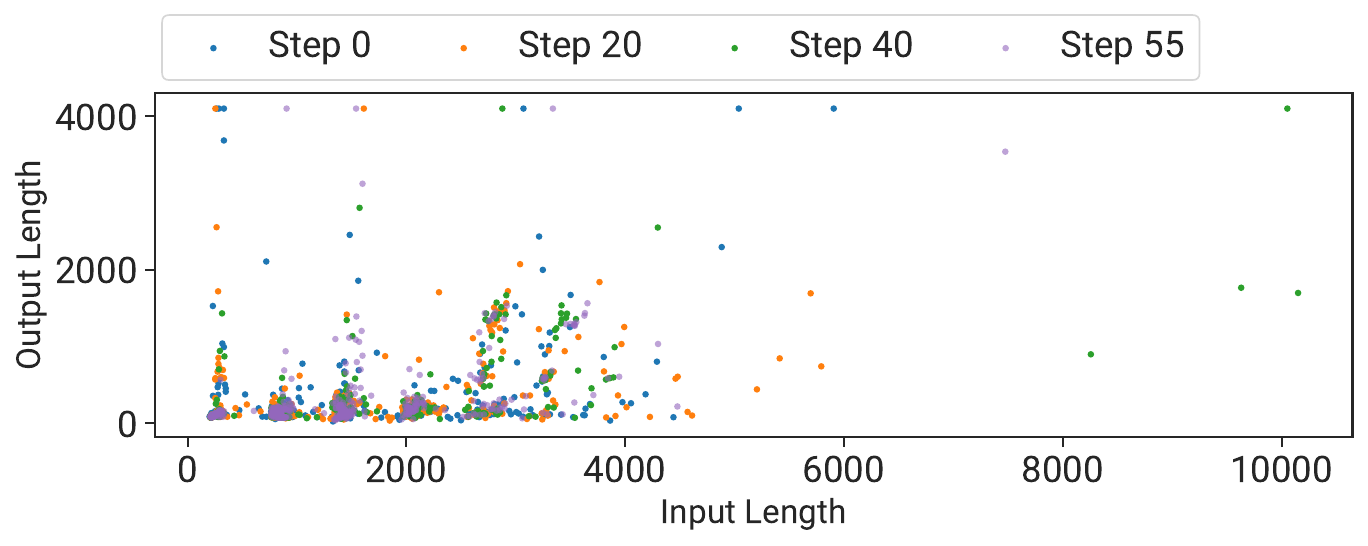}
    \caption{Scatter of \emph{Tool-use Task} input and output length joint distribution}
    \label{tool_use_input_output_scatter}
\end{figure}

\begin{figure}[t]
    \centering
    \includegraphics[width=\linewidth]{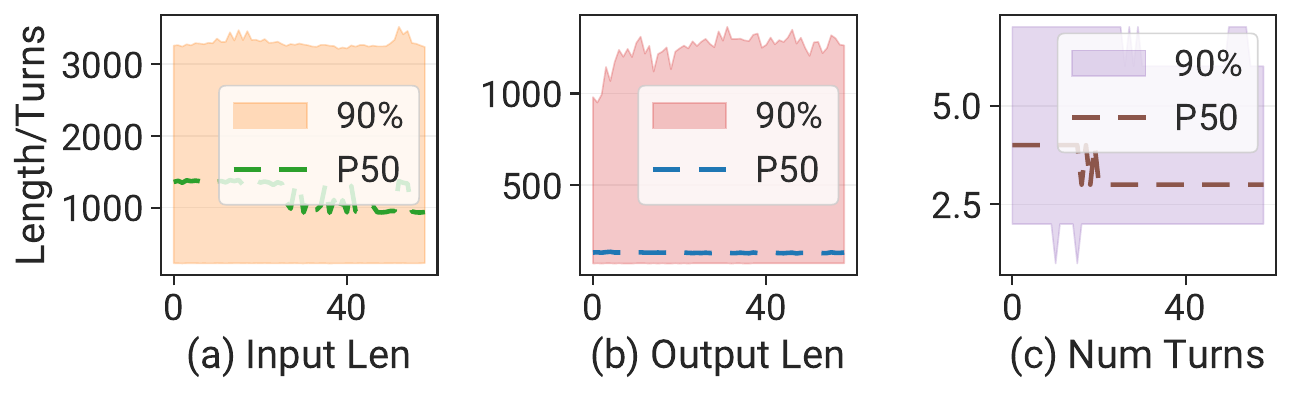}
    \caption{The workloads distribution across training steps of \emph{Tool-use Task}.}
    \label{tooluse_step_distribution}
\end{figure}

\begin{figure}[t]
    \centering
    \includegraphics[width=\linewidth]{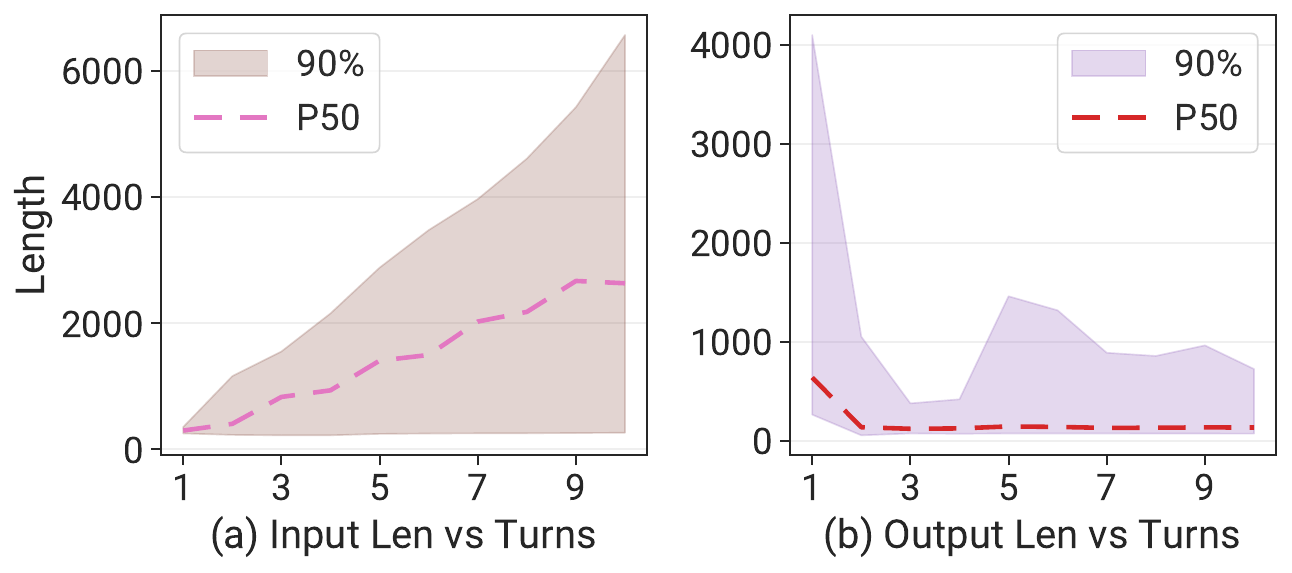}
    \caption{The workloads distribution across training steps of \emph{Tool-use Task}.}
    \label{tooluse_turns_distribution}
\end{figure}

Beyond the mathematics task, we further conduct a comprehensive analysis of the tool-use task in Model B.

\textbf{Highly skewed input/output length distributions.}
Figure \ref{tool_use_input_output_cdf} presents the input/output length distributions and the distribution of dialogue turns for the tool-use task. Compared with the previous two tasks, both input and output lengths exhibit pronounced long-tail behavior. For example, the 90th percentile of input length is 2{,}802 tokens, yet the maximum reaches 50k (about 18$\times$). This arises from two factors: (i) the long tail in the number of dialogue turns (Figure \ref{tool_use_input_output_cdf} (c)), where, in multi-turn settings, prior inputs and outputs are concatenated into the current turn’s input—more turns imply longer inputs; and (ii) in extreme cases, tool-return payloads can be very large (20–30k tokens), producing a markedly heavy-tailed input distribution (Figure \ref{tool_use_input_output_cdf} (d)). On the output side, the 90th percentile is 513 tokens, and the maximum exceeds 4k. Outputs are much shorter than in the two tasks we discussed before, because tool-centric RL typically follows a short thought chain—\emph{think → call tool(s) → analyze return → conclude}—rather than producing long chain-of-thought reasoning. As shown in Figure \ref{tool_use_input_output_cdf}, unlike the traces in LLM serving\cite{mooncake}, the turn count itself does not exhibit an extremely long tail as in reasoning traces: the maximum is only 10 turns, and roughly 80\% of dialogues fall between 3 and 6 turns. Moreover, the tool return length is concentrated between 470 and 490. It is because tool calls return an image in most cases, which is then processed through a vision encoder to obtain the same length. Overall, tool-use RL shows more complex distributional characteristics than the other tasks, posing additional challenges for RL system design and optimization.

\textbf{Clustered joint input–output distributions.}
In Figure \ref{tool_use_input_output_scatter}, we analyze the joint distribution of input and output lengths for the tool-use task. Unlike the largely uncorrelated patterns observed in the previous two tasks, the joint distribution here clusters by the number of dialogue turns, forming roughly five groups, with a few isolated outliers around the cluster peripheries. This corresponds to the concentration of turns between 2\textasciitilde6 and most samples share similar length profiles in Figure \ref{tool_use_input_output_cdf}. However, the presence of isolated outliers complicates RL training system optimizations.

\textbf{Turn-related input/output length distributions.}
We further study how length distributions relate to dialogue turns and training steps. In Figure \ref{tooluse_turns_distribution} (a), input length scales almost linearly with the number of turns, because each additional turn prepends prior turns’ inputs and outputs to the LLM context, and per-turn reasoning chains and tool-return lengths are themselves relatively concentrated (see Figure \ref{tool_use_input_output_cdf} (b) and Figure \ref{tool_use_input_output_cdf} (d)). In Figure \ref{tooluse_turns_distribution} (b). Interestingly, output length also depends on turn count: most long-tail outputs occur at a single turn, suggesting that, absent tool calls, the model may produce longer chains of thought. Once tool calls are involved, output length drops markedly, with the median decreasing from about 639 tokens to roughly 130 tokens, indicating that the model can quickly integrate context and make a tool-calling decision with far fewer tokens.

\textbf{Step-related input length dynamics.}
We next analyze how length distributions evolve with training steps in the tool-use task. Unlike the previous two tasks, the input length exhibits a distinct regime shift. As shown in Figure \ref{tooluse_step_distribution} (a), there is a sudden drop around step~30, after which the series transitions from stable to sustained fluctuations. The reason shown in Figure \ref{tooluse_step_distribution} (c), as training progresses, tool calls become more efficient and less frequent, reducing the number of dialogue turns and thus the total input length. Figure \ref{tooluse_step_distribution} (b) shows that output lengths remain nearly unchanged throughout training, implying a stable deliberation process when the model decides whether and how to invoke tools.


\subsection{Other-task Workloads}\label{workload_distribution}

\begin{figure}[t]
    \centering
    \includegraphics[width=1\linewidth]{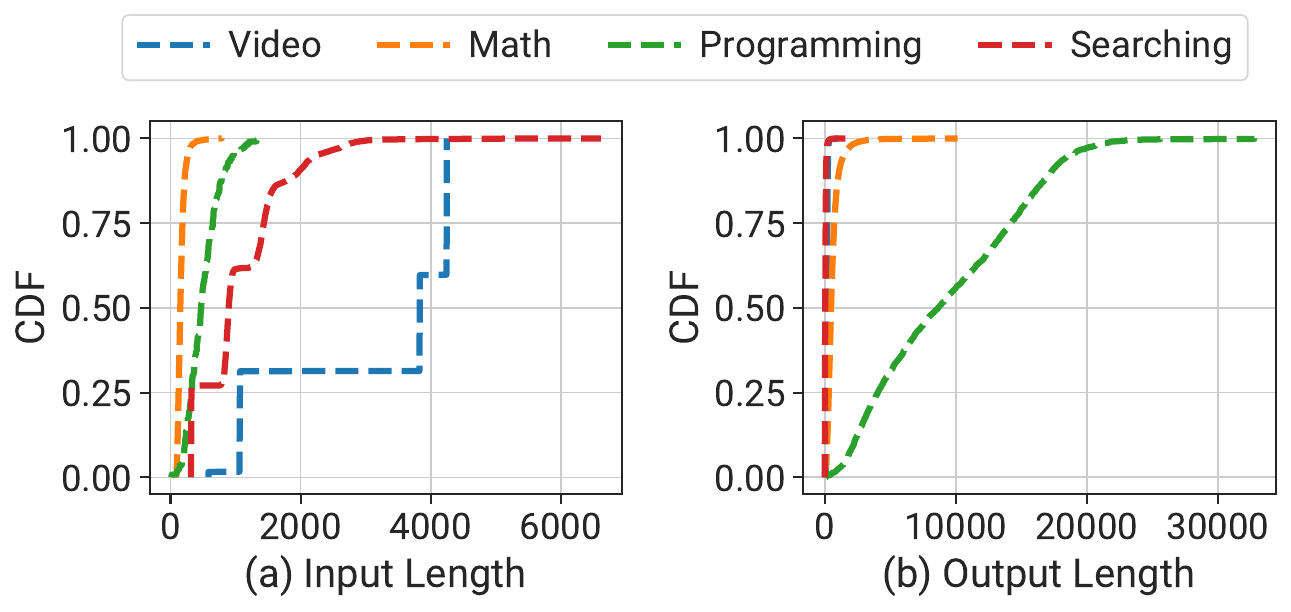}
    \caption{Cumulative distribution function (CDF) plots of the input and output length for each task.}
    \label{figure_cdf}
\end{figure}

\begin{figure}[t]
    \centering
    \includegraphics[width=1\linewidth]{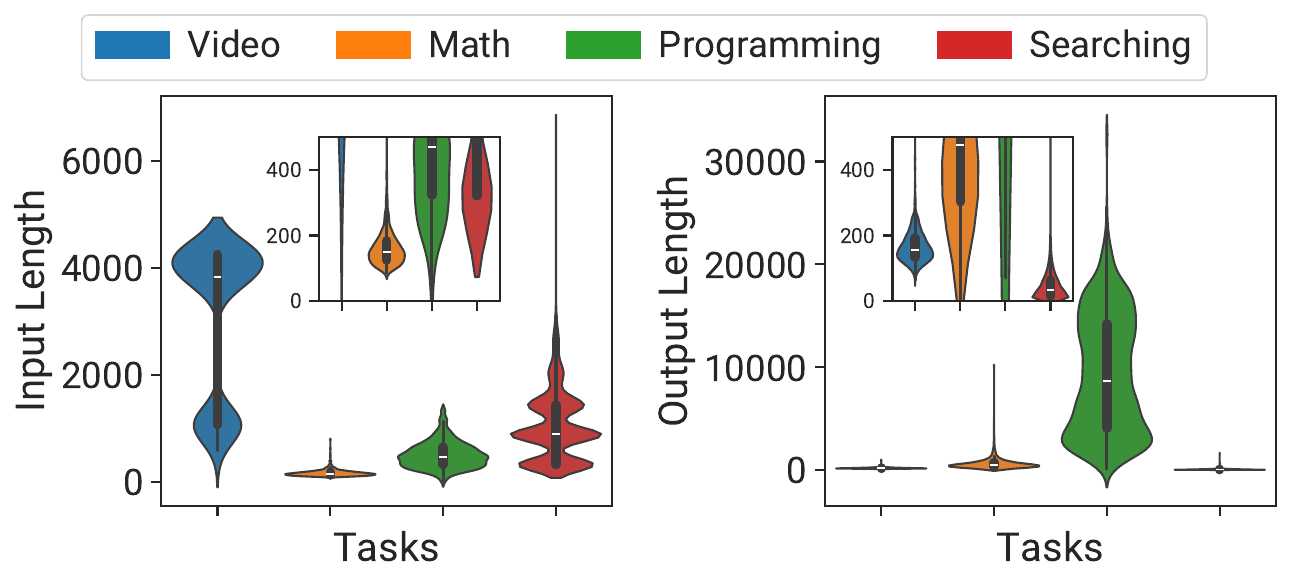}
    \caption{Violin plots of the input and output length for each task.}
    \label{figure_violin}
\end{figure}

\begin{figure}[t]
    \centering
    \includegraphics[width=\linewidth]{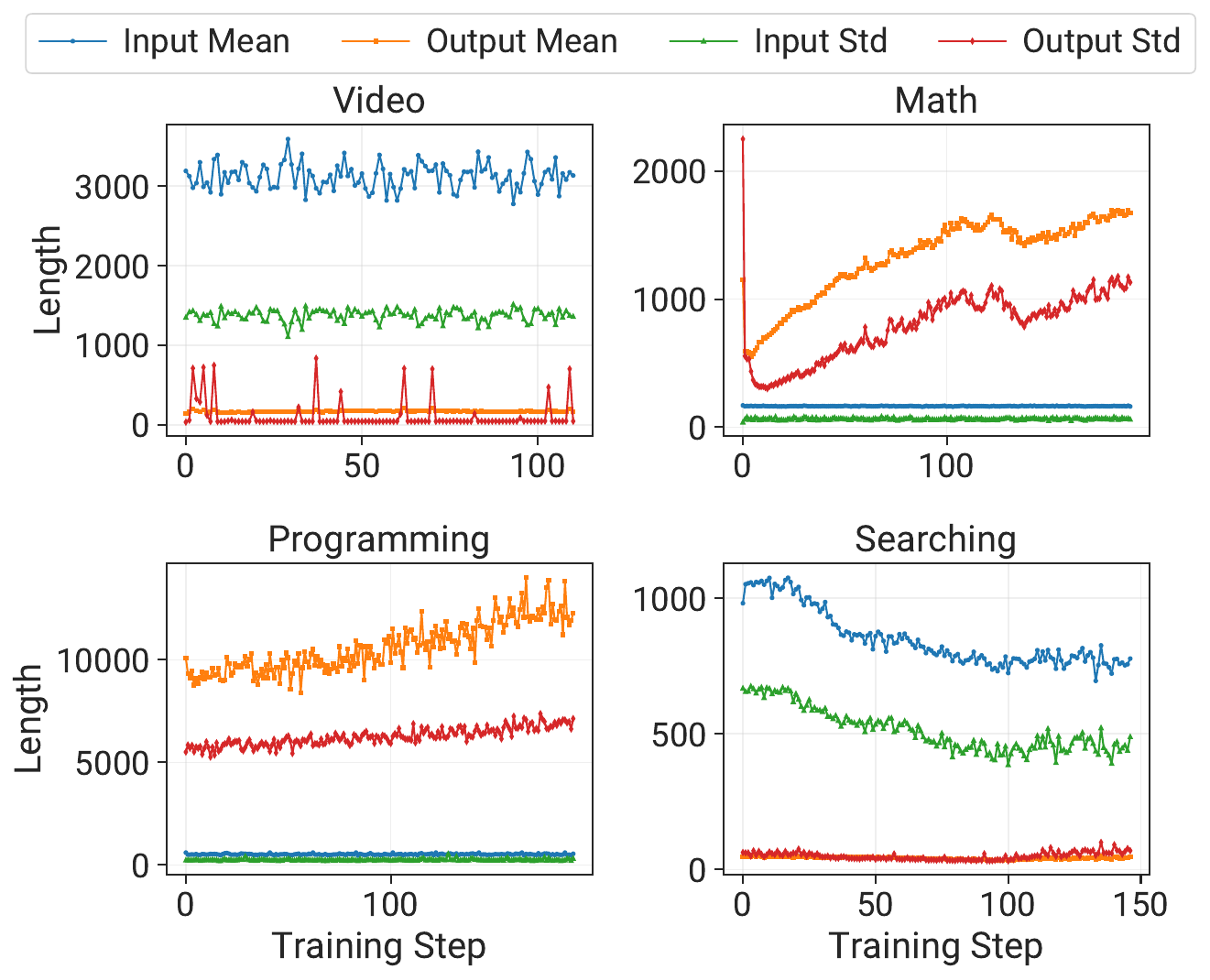}
    \caption{The trend of workload variation among different tasks.}
    \label{figure_temporal_trends}
\end{figure}

To further analyze the characteristics of workload distribution in RL training, we collect and analyze the workloads of four RL tasks with relatively smaller models.

\textbf{Various sequence length distribution among RL tasks.}
As shown in Figure \ref{figure_cdf}, within a single iteration, the distributions of different RL training tasks exhibit distinct characteristics. In terms of output length, programming tasks produce very long outputs, with an average of 9,329, whereas the video understanding and searching tasks average only 166.5 and 48.7, respectively. The pattern is reversed for input lengths: due to the presence of video information, the video understanding task has an average input length of 3,123.1, while the other tasks do not exceed 1k. Looking at the standard deviation of outputs, the coding task is the most uneven, with a standard deviation reaching 8,651.5, followed by the mathematics task at 595. By contrast, the multimodal task shows the most uneven input distribution, mainly because video length varies. As shown in Figures \ref{figure_cdf} and \ref{figure_violin}, both the mathematics and programming tasks display pronounced heavy-tailed behavior in their output distributions, whereas the other two tasks do not. In contrast, the searching task exhibits a pronounced heavy tail in its input-length distribution, which is not evident for the other three tasks. This is primarily because searching tasks require multiple external tool calls to retrieve information and passing it to the model, the number of tool invocations varies, leading to long and highly uneven inputs. We emphasize the need to comprehensively consider these workloads when optimizing RL training systems.


\textbf{Markedly different sequence length variation patterns. }
Compared to LLM RL training, the input and output length evolve continuously in relatively smaller models, with both the trend and magnitude of change differing across task types as shown in Figure \ref{figure_temporal_trends}. For example, in mathematics and programming tasks, output length tends to increase over time. In agent tasks, output length gradually decreases, whereas in the video understanding task, the input and output lengths tend to stabilize. This phenomenon arises because different RL tasks prioritize distinct model capabilities. For instance, coding and mathematics tasks emphasize strengthening reasoning, which encourages the model to produce longer chains of thought; accordingly, output lengths tend to increase as training progresses. In contrast, the searching task focuses on decision-making proficiency. As training advances, the model becomes more adept at tool use, thereby shortening the chain of thought required to complete a problem. Compared with large-scale models in Section \ref{sec_characterization}, small models exhibit significantly greater workload variability. We highlight the necessity of workload awareness when performing dynamic parallelism strategy selection in RL system optimization.

The widespread use of mixed datasets to boost overall capability further increases distributional diversity\cite{kimi_vl, seed1_5}, adding to the challenges of RL framework design and optimization.
\section{Fine-grained RL Training System Analysis}
\label{sec_profile}
This section analyzes system implications of RL training tasks as introduced in Section \ref{repre_algo} (the first four tasks). Firstly, we provide a time breakdown analysis of the different components of RL tasks in Section \ref{time_breakdown}, including fine-grained and coarse-grained. Moreover, we analyze the GPU memory usage in Section \ref{GPU_memory}. Lastly, we conduct a scalability study (Section \ref{scale_study}) and hyper hyperparameter study (Section \ref{hyper_parameter_study}) across RL tasks.

\subsection{Experiment Setup}

\textbf{Algorithm}: The detailed configurations of the four open source algorithms we analyze are presented in Section \ref{repre_algo}.\newline \textbf{Hardware configuration}: Our experiments are conducted on two clusters. Cluster A comprises 16 machines (128 GPUs), where each machine is equipped with 8 NVIDIA A800 80GB GPUs interconnected via 400GB/s NVLink, with an inter-machine bandwidth of 200Gbps. Cluster B consists of 2 machines, each equipped with 8 NVIDIA H800 80GB GPUs interconnected via 400GB/s NVLink, with an inter-node bandwidth of 1600Gbps. Our experiments utilize the following software versions: CUDA 12.4\cite{nvidia2024cuda}, PyTorch 2.6.0, NCCL 2.21.5\cite{nccl}, Verl 0.3.1, Sglang 0.4.6 and vLLM 0.8.5. 


\subsection{Time Breakdown}
\label{time_breakdown}
\begin{figure}[t]
    \centering
    \includegraphics[width=1\linewidth]{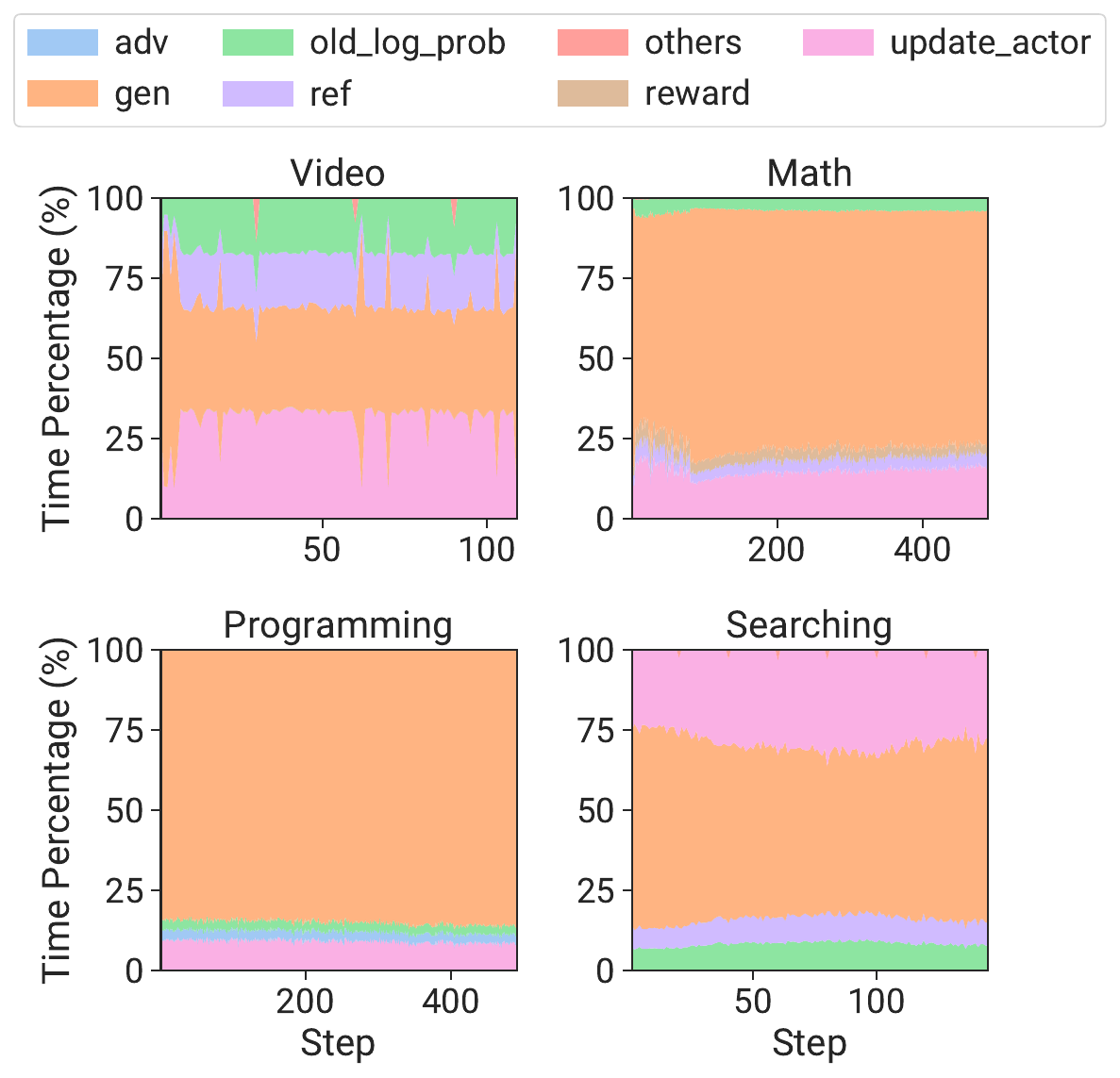}
    \caption{Coarse-grained time breakdown across different RL tasks.}
    \label{coarse_grained_time_breakdown}
\end{figure}

\begin{figure}[t]
    \centering
    \includegraphics[width=1\linewidth]{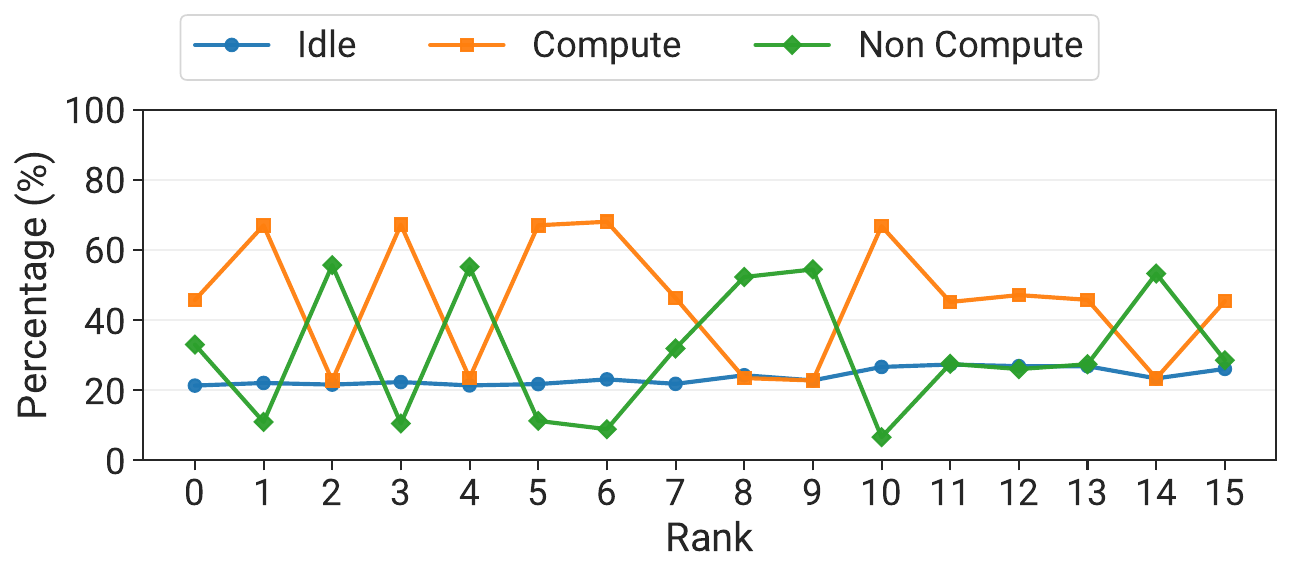}
    \caption{Per-rank time breakdown of video understanding task in training stage.}
    \label{perrank_time_breakdown}
\end{figure}

The workloads of different tasks vary significantly according to \TN, which directly affects the time breakdown. Analyzing these divergent breakdowns can provide guidance for optimizing bottlenecks in the RL training system.

\textbf{Coarse-grained time breakdown.} Stage-wise time shares in RL training are task-dependent. The rollout stage typically takes the lion’s share, and the breakdown fluctuates to varying extents. As shown in Figure \ref{coarse_grained_time_breakdown}, we observe substantial cross-task variation in the percentage of total time attributable to each stage. In most tasks, the rollout stage dominates the overall training runtime—for coding tasks, it accounts for over 80\%—whereas in some video understanding tasks it accounts for only about 30-40\%, a substantial part of this is attributable to the sequence length distribution discussed above. Because the load distribution shifts during training, the time shares of RL stages can be dynamic, especially in the video understanding task. For example, at step 62, there is a sharp spike in rollout time that directly alters the overall stage-wise breakdown. This spike arises from the sudden occurrence of long-tail samples, which induce fluctuations in rollout time. According to Amdahl’s law\cite{amdalhl}, we should prioritize optimizing the rollout phase. However, the complexity and diversity of the distributions simultaneously pose significant challenges, and also provide opportunities.

\begin{figure}[t]
    \centering
    \includegraphics[width=1\linewidth]{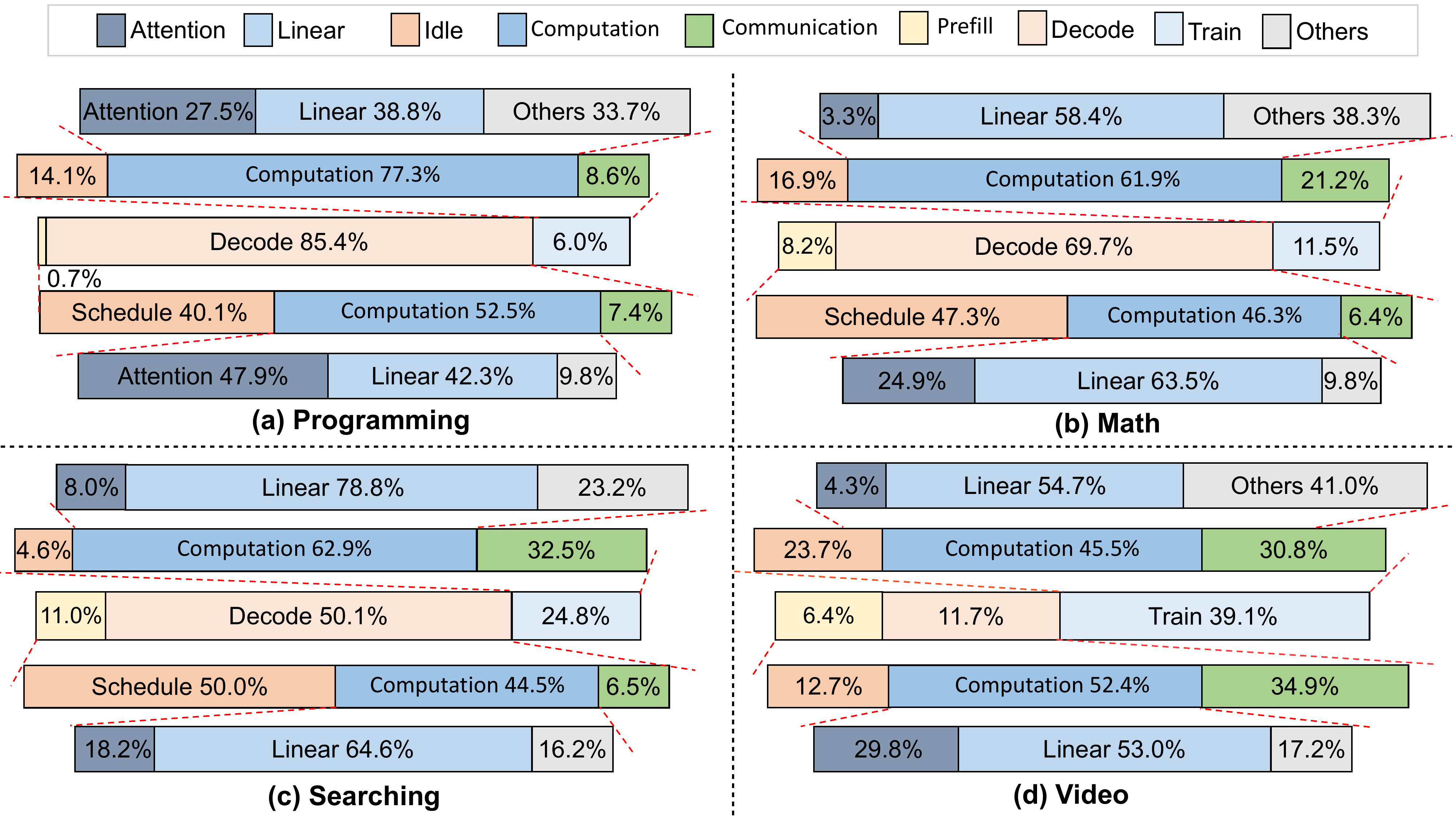}
    \caption{Fine-grained time breakdown across different RL tasks. We conduct a detailed profiling of the four tasks on Cluster B, for readability, we focus our in-depth analysis on the rollout and training phases, which typically dominate the overall training time; moreover, the computational characteristics of the inference phase are similar to those of training.}
    \label{fine_grained_time_breakdown}
\end{figure}

\textbf{Fine-grained time breakdown.} As shown in Figure \ref{fine_grained_time_breakdown}, we perform a finer-grained analysis of the two primary stages—rollout and training. The results reveal substantial cross-task differences in runtime composition, reflecting the underlying sequence length distributions. For the programming task, for instance, decoding accounts for 85.4\% of the time, whereas it is only 11.7\% for the video understanding task. This disparity directly correlates with the output-length distribution; longer outputs yield a larger decoding share. For tasks with long input sequences (e.g., video understanding), the rollout phase incurs higher communication overhead, primarily due to greater activation exchange of tensor parallel in the prefill phase.

Moreover, we observe considerable scheduling overhead during rollout. The root cause is that contemporary RL frameworks commonly employ inference engines as rollout backend; these engines are designed for online serving scenarios such as iteration-level scheduling and memory management, but these strategies introduce significant CPU overhead. Underscore the need for low CPU overhead inference backends.

In the training stage, we likewise observe substantial communication overhead. For example, in the video understanding task, communication accounts for 32.5\% of training time. Two factors primarily contribute: (i) a suboptimal parallelization strategy (here, a simple FSDP setup) that fails to sufficiently hide communication behind computation; and (ii) waiting delays in collective operations due to load imbalance, as illustrated in Figure \ref{perrank_time_breakdown}. In the multimodal understanding task, heterogeneity in image and video sizes further exacerbates compute imbalance, indicating that sample-level load-balancing schemes are too coarse-grained.

\begin{figure}[t]
    \centering
    \includegraphics[width=1\linewidth]{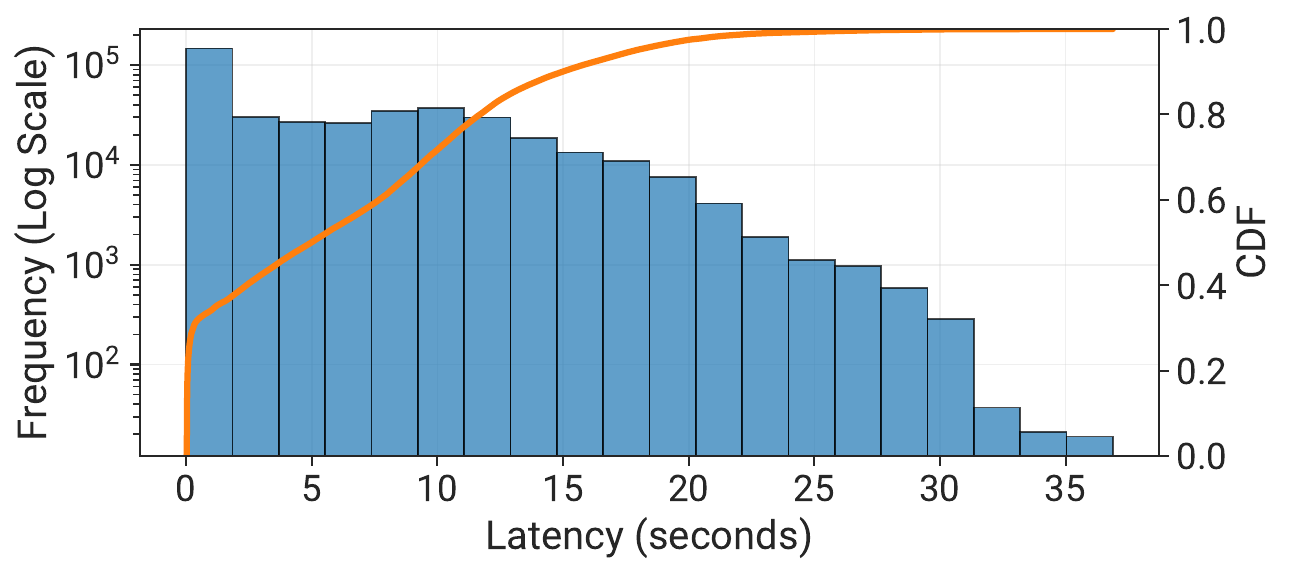}
    \caption{Histogram and CDF search tool calling latency of Searching Task.}
    \label{tool_calling_time_analysis}
\end{figure}

\textbf{Tool calling latency is likewise unstable.} As illustrated in Figure \ref{tool_calling_time_analysis}. In a searching task, the searching cost correlates with the difficulty of the query varies a lot. In tool call tasks, some invocations run locally (e.g., calling a Python function, local content retrieval) and are relatively stable, but many rely on remote services (e.g., E2B\cite{e2b}, AI applications\cite{mcp}, model endpoints), whose latency is affected by complex conditions such as network latency, making them inherently more volatile. From the perspective of the RL training framework, asynchronous tool invocation can hide tool latency, and employing caching can mitigate the impact of network instability. Additionally, these remote services entail considerable costs, rendering cost reduction \cite{ZeroSearch} in RL training a compelling area for investigation.




\textbf{Inefficient data transmission and management implementation.}
Contemporary RL frameworks often adopt a single controller design\cite{verl}, whereby all inter-stage data transfers must traverse the controller node. This mainly leads to 2 issues: first, in multimodal and long-context tasks, data volumes can be orders of magnitude larger than in standard tasks, making collection and dispatch via the controller prohibitively expensive. For example, in our measurements in cluster A on the current Verl implementation, Ray's costly serialization/deserialization yields only approximately 0.5 GB/s effective bandwidth within a node and roughly 0.015 GB/s across nodes. Second, funneling all data through a single controller can cause CPU out-of-memory (OOM) on a single node (Sec. \ref{hyper_parameter_study}), resulting in training failures. There is a pressing need for a more efficient mechanism to transfer and manage inter-stage data in RL training.

\textbf{Suboptimal parameter synchronization mechanism.} Parameter synchronization further involves cross-framework interaction and becomes increasingly complex and time-consuming in large model training, where there remains room for optimization. Mainstream open-source approaches include HybridEngine, from-disk updates, and all-gather plus broadcast. HybridEngine\cite{verl} co-locates rollout and training, simplifying parameter synchronization. The from-disk approach first persists updated parameters, after which the inference engine reloads them from the file system. It is really slow but 
suitable for auto scaling training. The all-gather plus broadcast scheme first all-gathers parameters to rank 0, which then broadcasts them to the inference engine's tensor-parallel (TP) group\cite{openrlhf}. These approaches either incur redundant communication or lack generality, highlighting the need for a unified, efficient parameter synchronization mechanism.



\subsection{GPU Memory Footprint}
\label{GPU_memory}

\begin{figure}[t]
    \centering
    \includegraphics[width=1\linewidth]{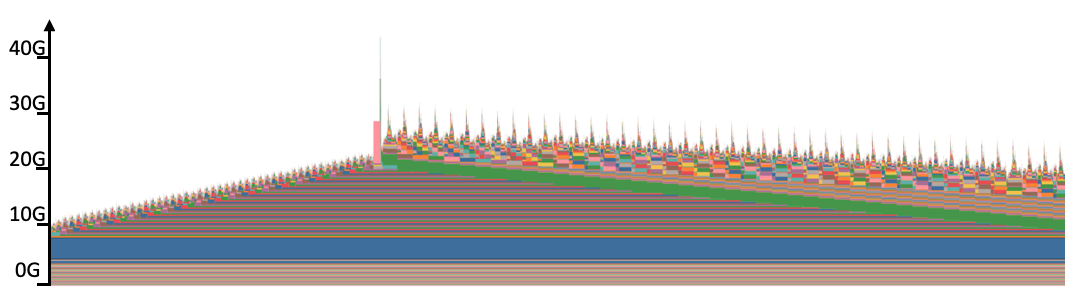}
    \caption{Memory snapshot\cite{torch-memory-snapshot} of DeepCoder's training stage. Since each task exhibits similar GPU memory usage during the training stage, we use the programming task for illustration. We offload optimizer states to host memory to prevent OOM.}
    \label{memory_footprint}
\end{figure}

During RL training, multiple model parameters are involved. For models that are being trained (e.g., the actor model), additional storage must be allocated for gradients and optimizer states\cite{adam}. Furthermore, for models that run on the inference engine, sufficient KV-cache storage must also be provisioned\cite{vllm}. These memory consumers do not coexist throughout the entire process, posing both spatial and temporal challenges for memory management.

\textbf{Static, coarse-grained memory optimization during training.} In RL training, conservative GPU memory policies are often adopted to prevent OOM: static memory footprints (e.g., optimizer states) are preferentially offloaded to host memory, and parameters and gradients are offloaded when device memory becomes insufficient\cite{Zero, Zero_offload}. In addition, as indicated by certain spikes in Figure \ref{memory_footprint}, the long sequence lengths during training lead to substantial activation footprints\cite{li2021sequence, jacobs2023deepspeed}, for which recomputation\cite{gpipe} is commonly employed. Such coarse-grained, static strategies result in suboptimal GPU memory utilization and extra computation or transfer overheads, thereby degrading RL training efficiency.

\textbf{Load-unaware KV-cache preemption policy.} At scales of 16 GPUs, the programming task exhibits frequent KV-cache preemption with \texttt{tensor parallelism} = 8. This arises because the task generally produces long sequences, demanding substantial GPU memory for KV-cache storage. These preemptions incur KV-cache recomputation and directly degrade training efficiency. While existing KV-cache preemption optimizations target inference serving\cite{kvcache_preemption_1, kvcache_preemption_2}, RL training differs substantially and remains largely unexplored.

\subsection{Scale Study}
\label{scale_study}

\begin{figure}[t]
    \centering
    \includegraphics[width=1\linewidth]{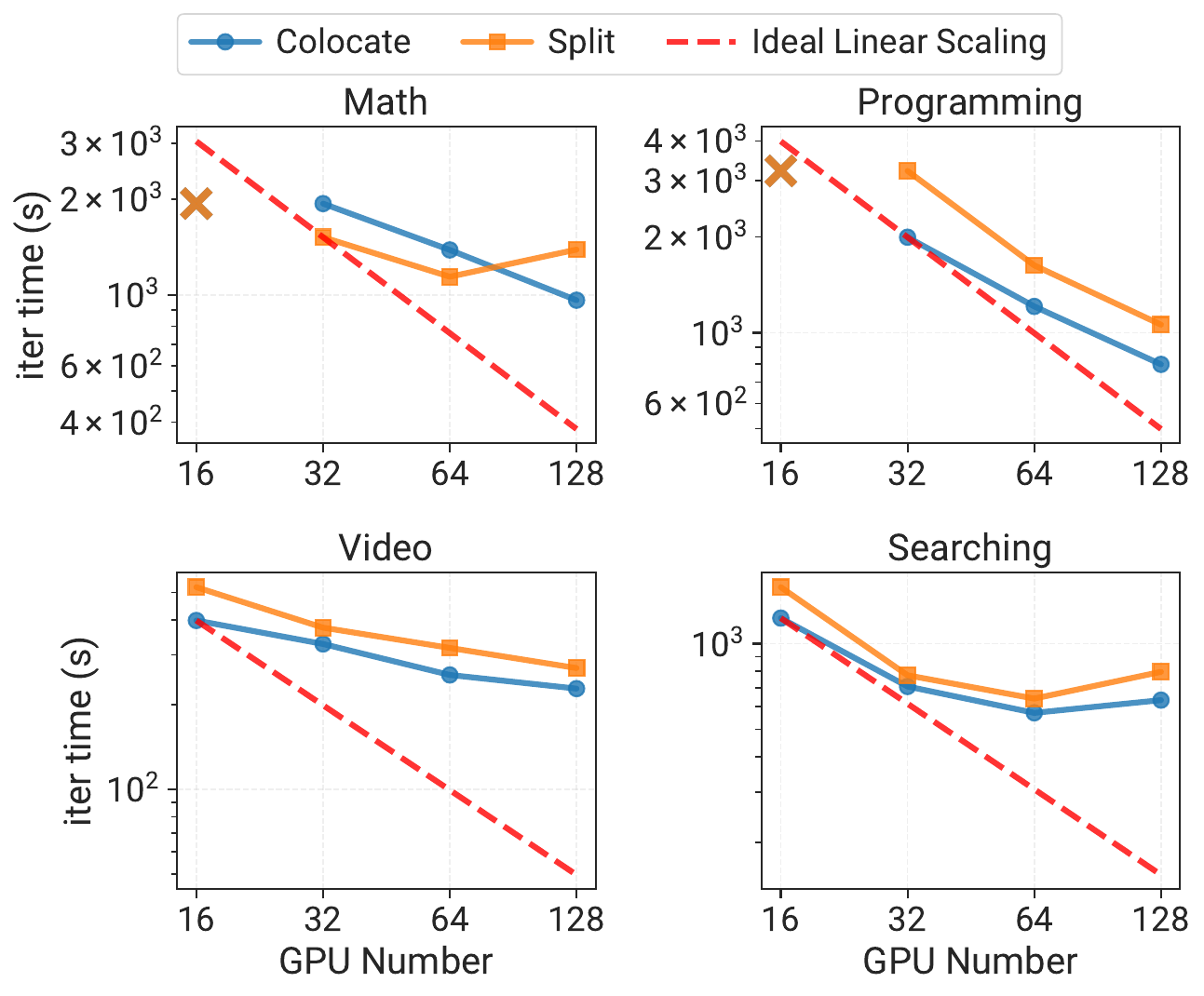}
    \caption{Scalability performance across different RL tasks. Experiments were conducted on cluster A. Split refers to dividing the GPU resources evenly into two parts: one part is allocated to the rollout phase, and the other to the inference and training phases.}
    \label{scalability}
\end{figure}

\begin{figure}[t]
    \centering
    \includegraphics[width=1\linewidth]{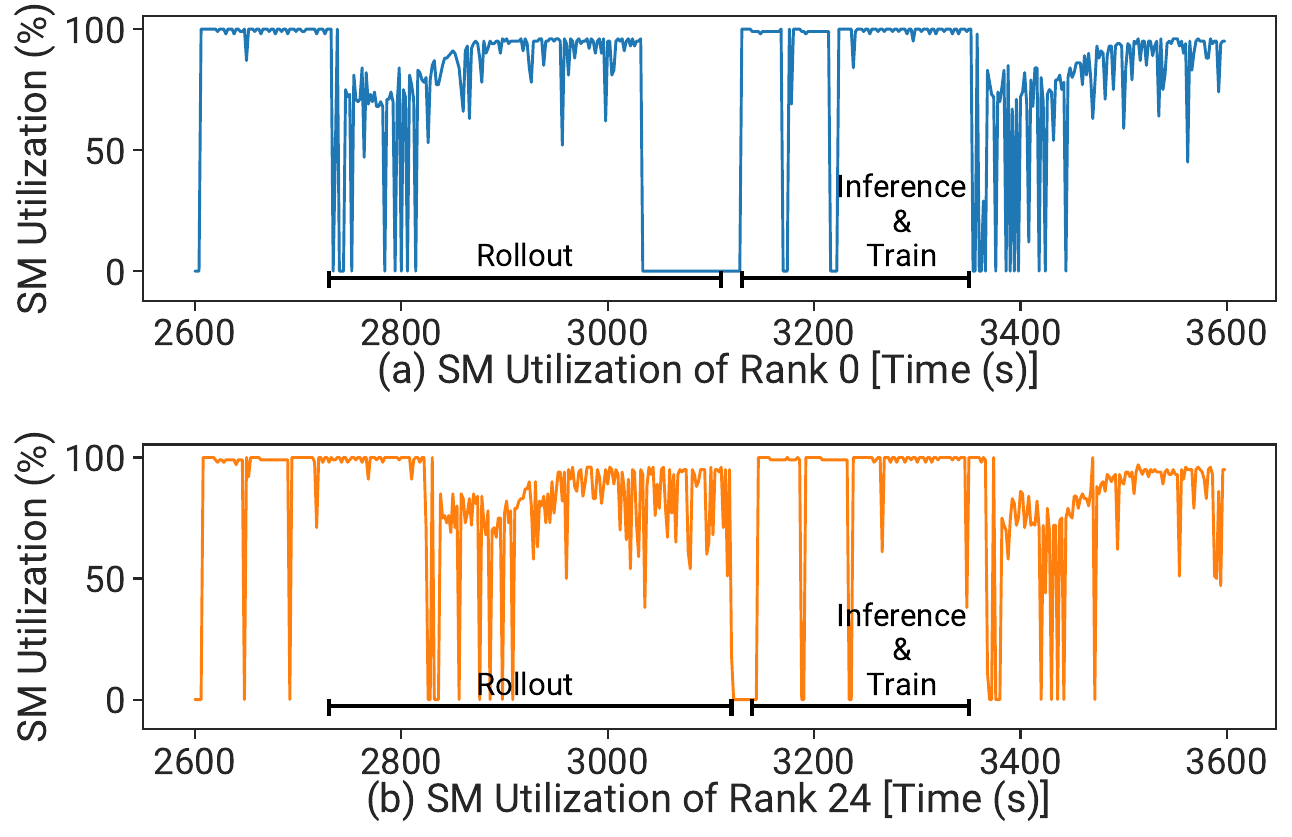}
    \caption{GPU SM utilization of RL training a 32B LLM using the DAPO algorithm over 32 GPUs.}
    \label{sm_util}
\end{figure}

In this section, we examine the strong scalability of the mainstream RL training framework across different tasks. As shown in Figure \ref{scalability}, we find that different tasks exhibit distinct strong-scaling behavior due to variations in model size and workload distribution. None of the tasks exhibits strong scalability; some tasks (e.g., mathematics) even display negative scaling. To provide an explanation for this observation, we outline four potential factors:

(1) \emph{Large GPU idling during rollout.}
As analyzed in prior sections, the long-tail output phenomenon, where a handful of extremely long sequences appear only on a few inference ranks, forces the remaining ranks to idle until they finish. As depicted in the Figure \ref{sm_util}, in the latter portion of the rollout phase, the GPU streaming multiprocessor (SM) utilization for rank 0 reaches 0, whereas rank 24 remains active.

(2) \emph{Poor GPU utilization in the rollout phase.}
During the 2730–3030,s interval shown in Figure \ref{sm_util}, the rollout stage is dominated by decoding, which fails to fully exploit GPU compute capacity. Moreover, the decode phase interleaves CPU scheduling, creating short but frequent GPU idles.

(3) \emph{High communication cost.}
As the cluster scales, the aggregate tensor-parallel (TP) communication overhead—especially in the prefill phase—grows and remains uncompensated by overlap. In the searching task rollout stage, 128 GPUs require 585.54,s versus 492.26,s for 64 GPUs (Figure \ref{scalability}). An analogous reversal is observed in training: the math task completes in 301.35,s with 32 GPUs, yet increases to 448.93,s when scaled to 128 GPUs. The reason is that the training stage becomes communication-bound at scale.

\subsection{Hyper Parameters Study}
\label{hyper_parameter_study}
Due to the importance of hyperparameters in system optimization\cite{Hydro,Pollux}, in this section, we examine hyper parameters affection to RL training system throughput. Specifically, we discuss the effects of \emph{batch size}, \emph{maximum response length}, and, in asynchronous RL training, \emph{the maximum permitted staleness} on training throughput.

\begin{figure}[t]
    \centering
    \includegraphics[width=1\linewidth]{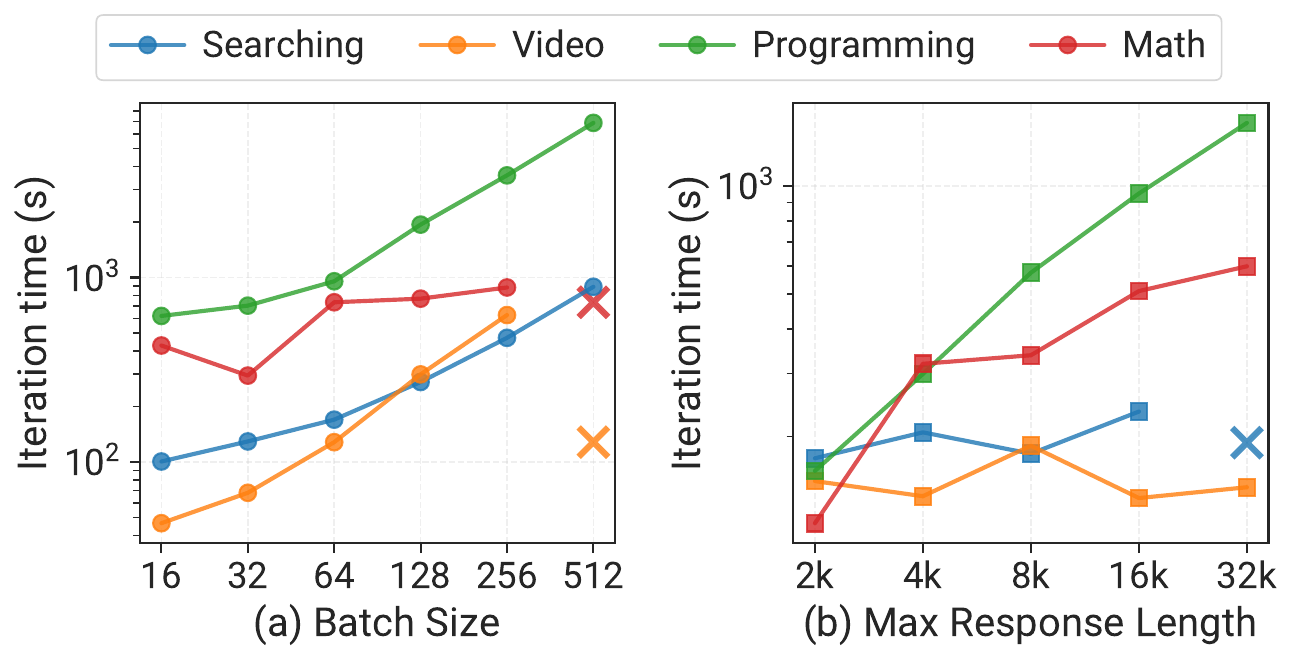}
    \caption{The effect of hyperparameters on different RL tasks. Experiment conducted on cluster B. In Figure (a), execution fails when the batch size is 512 due to a CPU OOM error, while in Figure (b), the failure is caused by the output length exceeding the model's maximum context length.}
    \label{hyperparameter_influence}
\end{figure}

\textbf{Batch size (bsz).} In LLM training, batch size is often pivotal for system performance; larger batches can better utilize GPU compute, create opportunities to hide communication overhead, and scale to huge systems\cite{Efficient21}. In inference, larger \emph{bsz} also offers more scheduling opportunities\cite{BatchLLM,BlendServe}. However, as shown in the Figure \ref{hyperparameter_influence} (a), the impact of batch size on iteration time in RL training different tasks is affected by bsz in distinct ways, with the following characteristics:

(1) The sensitivity to \emph{bsz} varies across tasks and is strongly influenced by the model size and the input/output length distributions used by each task. For example, in the searching and video understanding tasks, even with comparable model sizes, differing input/output distribution characteristics lead to markedly different growth trends. In the math task, the extremely long-tailed distribution renders the effect of \emph{bsz} on iteration time irregular.

(2) At large \emph{bsz}, CPU out of memory emerges readily in math and video understanding tasks with different reasons. In math, offloading model parameters, gradients, and optimizer states to the CPU leads to excessive memory consumption. In video understanding task, the large bsz induces massive input data (e.g., videos), exhausting memory. This again highlights the importance of storage/memory management mechanisms in RL training.

(3) Notably, in the video understanding task, we observe that training throughput degrades as \emph{bsz} increases (iteration time increases from 116.86 s at bsz = 128 to 253.40 s (about 2.2×) at bsz = 256). This aligns with our earlier analysis of load imbalance in multimodal tasks. A larger \emph{bsz} implies more mini-batches, and each mini-batch can suffer imbalance, forcing collective operations to wait for the slowest rank.

\textbf{Maximum response length.} As shown in Figure \ref{hyperparameter_influence} (b), the impact of the maximum response length on iteration time varies substantially across tasks. The video understanding and searching tasks are almost unaffected by the maximum response length, whereas programming and math exhibit a clear positive correlation. These observations are consistent with our earlier analysis of sequence-length distributions. They also suggest optimization opportunities, such as selecting the maximum response length at the prompt level and adopting task-aware scheduling strategies.

\begin{table}[t]
\centering
\setlength{\tabcolsep}{4pt}       
\renewcommand{\arraystretch}{1.25} 
\resizebox{\linewidth}{!}{%
\begin{tabular}{lcccc}
\toprule
Maximum Permitted Staleness & Math500\cite{math500} & Aime24 & Aime23 & E2E Time \\
\midrule
0  & 84.4\% & 32.6\% & 74.3\% & 52h 14min \\
8  & 84.7\% & 32.5\% & 73.0\% & \textbf{33h 50min} \\
16 & 82.6\% & 32.7\% & 74.1\% & \textbf{35h 15min} \\
\bottomrule
\end{tabular}%
}
\caption{Results under different maximum permitted staleness. Conduct official DeepSeek-R1-Distill-Qwen-1.5B experiment running 80 training steps. For evaluation, we sample 32 responses per question, reporting the average pass@1 accuracy. Experiments are conducted on cluster B, one node for the rollout stage and one for the training stage.}
\label{asyncRL}
\end{table}

\textbf{Maximum Permitted Staleness.} As shown in Figure \ref{asyncRL}, asynchronous RL mitigates load imbalance in the rollout phase by placing rollout and training on different machines and executing them in a pipeline. As our analysis shows, rollout time often exceeds training time. Consequently, an asynchronous setup is commonly adopted in which the inference stack uses stale parameters that lag behind the training stage. The degree of staleness can be tuned via the maximum permitted staleness.

As shown in Table \ref{asyncRL}, asynchronous training delivers roughly a 60\% improvement in end-to-end (E2E) throughput over synchronous training, without degrading downstream task performance. Interestingly, the maximum permitted staleness values of 8 and 16 yield similar speedups. This is because, with two nodes, pipeline balance is achieved at around 1–2, so the upper bound is not the limiting factor.

Furthermore, as training scales up, deciding how to allocate GPU nodes between the rollout and training stages to achieve pipeline balance becomes a key systems question. Optimizing this allocation should explicitly account for the maximum permitted staleness in order to identify the optimal configuration. Moreover, the dynamism and heterogeneity of RL workloads also affect resource partitioning and call for adaptive rebalancing. Effective asynchronous RL training requires co-optimizing resource allocation with respect to cluster scale, the maximum permitted staleness, and the distribution of the workload.

\section{\TN Benchmark Suite}
\textbf{Motivation:} From a research perspective, the benchmark suite serves as a crucial tool for uncovering system insights (Section \ref{sec_profile}) and enabling researchers to validate their optimization methods effectively. Current RL training systems face challenges in new cases that are often difficult to identify and analyze without comprehensive workload characterization. \TN benchmark suite bridges this gap by providing diverse and realistic workloads from \TN. Also, with access to standardized workloads and performance baselines, researchers can focus on developing novel optimization strategies(e.g., dynamic parallel strategies\cite{Tenplex,Enabling24}) rather than spending extensive GPU hours collecting workload data, accelerating the pace of research innovation in RL training systems. From a system evaluation perspective, A benchmark suite provides a standardized platform for comprehensive assessment of RL training frameworks. The benchmark suite enables fair and consistent comparison across different RL training frameworks by providing standardized workloads and evaluation metrics, helping practitioners make decisions about framework selection.

\begin{figure}[t]
    \centering
    \includegraphics[width=\linewidth]{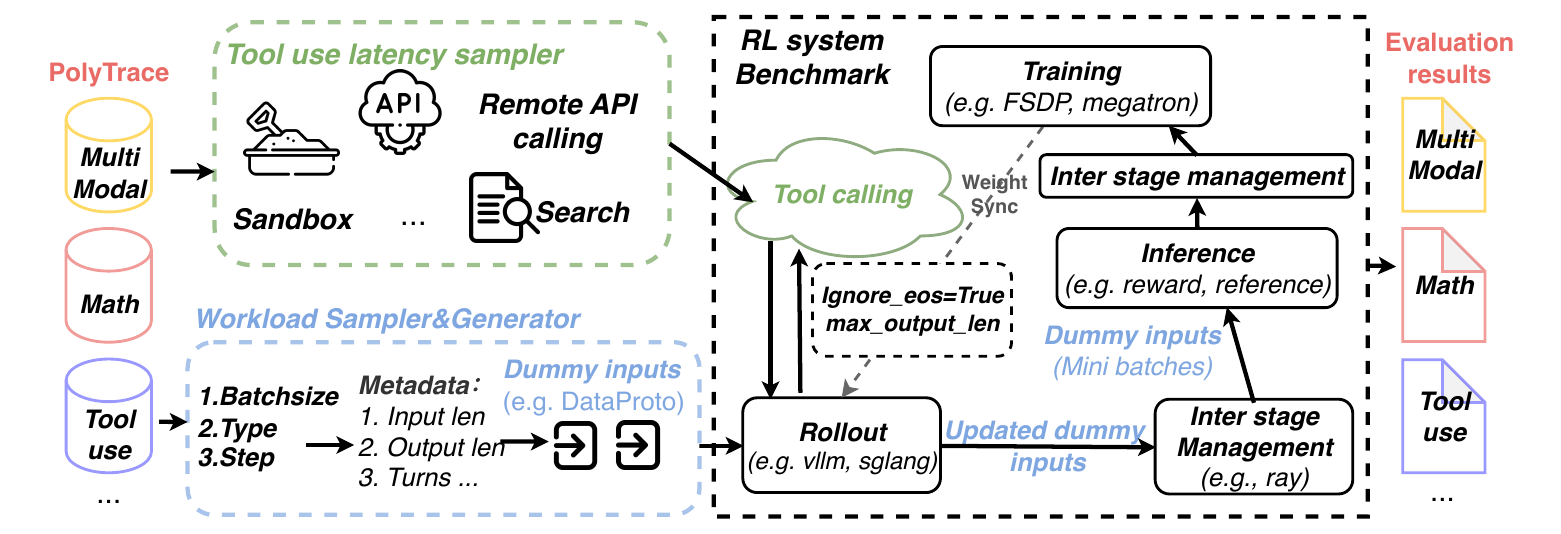}
    \caption{Benchmark Suite Overview.}
    \label{benchmark_suite}
\end{figure}

\noindent \textbf{Benchmark suite design}: Our benchmark suite can seamlessly integrate trace data into complex RL training frameworks, consisting of three main processes:

    \noindent \textbf{Tool use latency sampler}: Based on our previous analysis, the tool invocation time distribution in RL training is highly imbalanced. To simulate real-world scenarios, we first collect tool use times, then perform random sampling according to this latency distribution, enabling the rollout phase to avoid calling actual tools, which often incur significant deployment and invocation costs (e.g., Google's search API costs \$5 for 1000 calls). This allows us to obtain end-to-end performance of RL tasks in tool invocation scenarios.

    \noindent \textbf{Workload sampler and generator}: Based on our previous analysis, diverse and real-world workloads are critical for RL training system evaluation as they reveal different performance bottlenecks and ensure comprehensive assessment across realistic training scenarios. To generate workloads, we first sample the specified amount of metadata (including input data length, output data length, and number of multi-turn conversation rounds, etc.) from our trace data based on user-provided hyper parameters (including batch size, task type, step, etc.), then generate dummy inputs according to the data format specified in the RL training framework (such as DataProto in Verl) and put it to the RL training system.

    \noindent \textbf{RL system benchmark}: After receiving the dummy data generated by the workload sampler and generator, the RL training framework can begin executing the training process. In the rollout phase, we can control the output length of data by setting \texttt{ignore\_eos\_token=True} and \texttt{max\_output\_len=output length} for each piece of data, which is widely supported in current inference frameworks, to obtain the actual execution time. Then, the updated dummy data is collected through the inter-stage management backend (such as Ray\cite{Ray}) and undergoes optimization strategies such as load balancing\cite{verl} before being passed to the inference stage, where normal computational processes can proceed, ultimately yielding benchmark performance data.

\noindent \textbf{End-to-end accuracy}: To verify the accuracy of the benchmark suite, we use video understanding workloads as an example to conduct performance testing on the Verl framework. As shown in Figure \ref{fig:trace_record}, the input data format (\texttt{DataProto}) in the framework primarily consists of two components: \texttt{batch} and \texttt{no\_tensor\_batch}, requiring only the input/output length information and the shape of video data, which we obtain from \TN through the workload sampler. We then call random methods to populate the \texttt{DataProto} according to the specified format and data types, and directly input it into the Verl framework. The resulting end-to-end latency is 484.48s, while the actual latency is 511.44s, with an error of 5.3\%.
\section{Discussion}
\label{sec_discussion}



\textbf{Scope limitations.} Despite our best efforts to analyze LLM RL training systems and workloads, our study still has several limitations. Firstly, although our study covers multiple categories of RL tasks, the rapid evolution of RL algorithms means there are emerging task types we did not include. We believe the workloads we selected are representative and that the insights derived from our in-depth analysis are both valuable and broadly applicable. We leave the characterization of these newer workloads to future work. Secondly, our analysis primarily focuses on the RL framework. Other perspectives, such as fault tolerance, are left for future work. Nevertheless, we believe our methodology—characterizing real workloads in an emerging domain and examining how those workloads stress current systems—can be fruitfully applied to other LLM tasks and systems research.

\textbf{Continuous benchmark enhancement.} As RLVR continues to evolve, the benchmarks presented in this paper may be insufficient to fully capture the workloads of RLVF. In response, we will continually expand the diversity of benchmark tasks and incorporate realistic tool-invocation latencies to keep pace with rapidly advancing algorithms. Moreover, we are upgrading our benchmark suite by adding support for additional frameworks and introducing more fine-grained simulations, such as KV cache reuses. Furthermore, we are exploring promising directions, including the development of a simulator that does not require a real execution environment.







%
\section{Conclusion}
\label{sec_conclusion}

We present a practical, workload-centric study of RL for LLMs that combines a comprehensive survey with empirical profiling across diverse tasks. Our analysis exposes core challenges and bottlenecks in RL training, highlighting the outsized impact of workload characteristics on framework performance. By profiling three real-world, state-of-the-art RL workloads and augmenting them with open-source algorithmic tasks, we curate a workload dataset that both informs and motivates researchers—and serves as a principled benchmark for evaluating RL framework efficacy.


\bibliographystyle{plain}
\bibliography{arxiv}

\begin{thebibliography}{10}

\bibitem{amdalhl}
Gene~M Amdahl.
\newblock Validity of the single processor approach to achieving large scale computing capabilities.
\newblock In {\em Proceedings of the April 18-20, 1967, spring joint computer conference}, pages 483--485, 1967.

\bibitem{cosmos}
Alisson Azzolini, Junjie Bai, Hannah Brandon, Jiaxin Cao, Prithvijit Chattopadhyay, Huayu Chen, Jinju Chu, Yin Cui, Jenna Diamond, Yifan Ding, et~al.
\newblock Cosmos-reason1: From physical common sense to embodied reasoning.
\newblock {\em arXiv preprint arXiv:2503.15558}, 2025.

\bibitem{sandbox_fusion}
Bytedance-Seed-Foundation-Code-Team, :, Yao Cheng, Jianfeng Chen, Jie Chen, Li~Chen, Liyu Chen, Wentao Chen, Zhengyu Chen, Shijie Geng, Aoyan Li, Bo~Li, Bowen Li, Linyi Li, Boyi Liu, Jiaheng Liu, Kaibo Liu, Qi~Liu, Shukai Liu, Siyao Liu, Tianyi Liu, Tingkai Liu, Yongfei Liu, Rui Long, Jing Mai, Guanghan Ning, Z.~Y. Peng, Kai Shen, Jiahao Su, Jing Su, Tao Sun, Yifan Sun, Yunzhe Tao, Guoyin Wang, Siwei Wang, Xuwu Wang, Yite Wang, Zihan Wang, Jinxiang Xia, Liang Xiang, Xia Xiao, Yongsheng Xiao, Chenguang Xi, Shulin Xin, Jingjing Xu, Shikun Xu, Hongxia Yang, Jack Yang, Yingxiang Yang, Jianbo Yuan, Jun Zhang, Yufeng Zhang, Yuyu Zhang, Shen Zheng, He~Zhu, and Ming Zhu.
\newblock Fullstack bench: Evaluating llms as full stack coders, 2025.

\bibitem{SFT2}
Wei-Lin Chiang, Zhuohan Li, Ziqing Lin, Ying Sheng, Zhanghao Wu, Hao Zhang, Lianmin Zheng, Siyuan Zhuang, Yonghao Zhuang, Joseph~E Gonzalez, et~al.
\newblock Vicuna: An open-source chatbot impressing gpt-4 with 90\%* chatgpt quality.
\newblock {\em See https://vicuna. lmsys. org (accessed 14 April 2023)}, 2(3):6, 2023.

\bibitem{deepseekai2025deepseekr1incentivizingreasoningcapability}
DeepSeek-AI.
\newblock Deepseek-r1: Incentivizing reasoning capability in llms via reinforcement learning, 2025.

\bibitem{e2b}
{E2B} -- {Code} {Execution} {Environment} for {AI}.
\newblock \url{https://e2b.dev/}.
\newblock Accessed: 2025-09-12.

\bibitem{retool}
Jiazhan Feng, Shijue Huang, Xingwei Qu, Ge~Zhang, Yujia Qin, Baoquan Zhong, Chengquan Jiang, Jinxin Chi, and Wanjun Zhong.
\newblock Retool: Reinforcement learning for strategic tool use in llms, 2025.

\bibitem{video_r1}
Kaituo Feng, Kaixiong Gong, Bohao Li, Zonghao Guo, Yibing Wang, Tianshuo Peng, Benyou Wang, and Xiangyu Yue.
\newblock Video-r1: Reinforcing video reasoning in mllms.
\newblock {\em arXiv preprint arXiv:2503.21776}, 2025.

\bibitem{gpt3}
Luciano Floridi and Massimo Chiriatti.
\newblock Gpt-3: Its nature, scope, limits, and consequences.
\newblock {\em Minds and machines}, 30(4):681--694, 2020.

\bibitem{areal}
Wei Fu, Jiaxuan Gao, Xujie Shen, Chen Zhu, Zhiyu Mei, Chuyi He, Shusheng Xu, Guo Wei, Jun Mei, Jiashu Wang, Tongkai Yang, Binhang Yuan, and Yi~Wu.
\newblock Areal: A large-scale asynchronous reinforcement learning system for language reasoning, 2025.

\bibitem{bytescale}
Hao Ge, Junda Feng, Qi~Huang, Fangcheng Fu, Xiaonan Nie, Lei Zuo, Haibin Lin, Bin Cui, and Xin Liu.
\newblock Bytescale: Efficient scaling of llm training with a 2048k context length on more than 12,000 gpus.
\newblock {\em CoRR}, abs/2502.21231, 2025.

\bibitem{Enabling24}
Hao Ge, Fangcheng Fu, Haoyang Li, Xuanyu Wang, Sheng Lin, Yujie Wang, Xiaonan Nie, Hailin Zhang, Xupeng Miao, and Bin Cui.
\newblock Enabling parallelism hot switching for efficient training of large language models.
\newblock In {\em Proceedings of the ACM SIGOPS 30th Symposium on Operating Systems Principles}, SOSP '24. Association for Computing Machinery, 2024.

\bibitem{seed1_5}
Dong Guo, Faming Wu, Feida Zhu, Fuxing Leng, Guang Shi, Haobin Chen, Haoqi Fan, Jian Wang, Jianyu Jiang, Jiawei Wang, et~al.
\newblock Seed1. 5-vl technical report.
\newblock {\em arXiv preprint arXiv:2505.07062}, 2025.

\bibitem{asyncflow}
Zhenyu Han, Ansheng You, Haibo Wang, Kui Luo, Guang Yang, Wenqi Shi, Menglong Chen, Sicheng Zhang, Zeshun Lan, Chunshi Deng, et~al.
\newblock Asyncflow: An asynchronous streaming rl framework for efficient llm post-training.
\newblock {\em arXiv preprint arXiv:2507.01663}, 2025.

\bibitem{openrlhf}
Jian Hu, Xibin Wu, Zilin Zhu, Weixun Wang, Dehao Zhang, Yu~Cao, et~al.
\newblock Openrlhf: An easy-to-use, scalable and high-performance rlhf framework.
\newblock {\em arXiv preprint arXiv:2405.11143}, 2024.

\bibitem{helios}
Qinghao Hu, Peng Sun, Shengen Yan, Yonggang Wen, and Tianwei Zhang.
\newblock Characterization and prediction of deep learning workloads in large-scale gpu datacenters.
\newblock In {\em Proceedings of the International Conference for High Performance Computing, Networking, Storage and Analysis}, pages 1--15, 2021.

\bibitem{acme}
Qinghao Hu, Zhisheng Ye, Zerui Wang, Guoteng Wang, Meng Zhang, Qiaoling Chen, Peng Sun, Dahua Lin, Xiaolin Wang, Yingwei Luo, et~al.
\newblock Characterization of large language model development in the datacenter.
\newblock In {\em 21st USENIX Symposium on Networked Systems Design and Implementation (NSDI 24)}, pages 709--729, 2024.

\bibitem{Hydro}
Qinghao Hu, Zhisheng Ye, Meng Zhang, Qiaoling Chen, Peng Sun, Yonggang Wen, and Tianwei Zhang.
\newblock Hydro: {Surrogate-Based} hyperparameter tuning service in datacenters.
\newblock In {\em 17th USENIX Symposium on Operating Systems Design and Implementation}, OSDI '23, pages 757--777. USENIX Association, 2023.

\bibitem{gpipe}
Yanping Huang, Youlong Cheng, Ankur Bapna, Orhan Firat, Dehao Chen, Mia Chen, HyoukJoong Lee, Jiquan Ngiam, Quoc~V Le, Yonghui Wu, et~al.
\newblock Gpipe: Efficient training of giant neural networks using pipeline parallelism.
\newblock {\em Advances in neural information processing systems}, 32, 2019.

\bibitem{jacobs2023deepspeed}
Sam~Ade Jacobs, Masahiro Tanaka, Chengming Zhang, Minjia Zhang, Shuaiwen~Leon Song, Samyam Rajbhandari, and Yuxiong He.
\newblock Deepspeed ulysses: System optimizations for enabling training of extreme long sequence transformer models.
\newblock {\em arXiv preprint arXiv:2309.14509}, 2023.

\bibitem{LiveCodeBench}
Naman Jain, King Han, Alex Gu, Wen-Ding Li, Fanjia Yan, Tianjun Zhang, Sida Wang, Armando Solar-Lezama, Koushik Sen, and Ion Stoica.
\newblock Livecodebench: Holistic and contamination free evaluation of large language models for code.
\newblock {\em CoRR}, abs/2403.07974, 2024.

\bibitem{Philly}
Myeongjae Jeon, Shivaram Venkataraman, Amar Phanishayee, Junjie Qian, Wencong Xiao, and Fan Yang.
\newblock Analysis of $\{$Large-Scale$\}$$\{$Multi-Tenant$\}$$\{$GPU$\}$ clusters for $\{$DNN$\}$ training workloads.
\newblock In {\em 2019 USENIX Annual Technical Conference (USENIX ATC 19)}, pages 947--960, 2019.

\bibitem{MegaScale}
Ziheng Jiang, Haibin Lin, Yinmin Zhong, Qi~Huang, Yangrui Chen, Zhi Zhang, Yanghua Peng, Xiang Li, Cong Xie, Shibiao Nong, et~al.
\newblock $\{$MegaScale$\}$: Scaling large language model training to more than 10,000 $\{$GPUs$\}$.
\newblock In {\em 21st USENIX Symposium on Networked Systems Design and Implementation (NSDI 24)}, pages 745--760, 2024.

\bibitem{search_r1}
Bowen Jin, Hansi Zeng, Zhenrui Yue, Jinsung Yoon, Sercan Arik, Dong Wang, Hamed Zamani, and Jiawei Han.
\newblock Search-r1: Training llms to reason and leverage search engines with reinforcement learning.
\newblock {\em arXiv preprint arXiv:2503.09516}, 2025.

\bibitem{adam}
Diederik~P Kingma and Jimmy Ba.
\newblock Adam: A method for stochastic optimization.
\newblock {\em arXiv preprint arXiv:1412.6980}, 2014.

\bibitem{inference_training_analyze}
Joyjit Kundu, Wenzhe Guo, Ali BanaGozar, Udari De~Alwis, Sourav Sengupta, Puneet Gupta, and Arindam Mallik.
\newblock Performance modeling and workload analysis of distributed large language model training and inference.
\newblock In {\em 2024 IEEE International Symposium on Workload Characterization (IISWC)}, pages 57--67, 2024.

\bibitem{nq}
Tom Kwiatkowski, Jennimaria Palomaki, Olivia Redfield, Michael Collins, Ankur Parikh, Chris Alberti, Danielle Epstein, Illia Polosukhin, Jacob Devlin, Kenton Lee, Kristina Toutanova, Llion Jones, Matthew Kelcey, Ming-Wei Chang, Andrew~M. Dai, Jakob Uszkoreit, Quoc Le, and Slav Petrov.
\newblock Natural questions: A benchmark for question answering research.
\newblock {\em Transactions of the Association for Computational Linguistics}, 7:452--466, 2019.

\bibitem{vllm}
Woosuk Kwon, Zhuohan Li, Siyuan Zhuang, Ying Sheng, Lianmin Zheng, Cody~Hao Yu, Joseph Gonzalez, Hao Zhang, and Ion Stoica.
\newblock Efficient memory management for large language model serving with pagedattention.
\newblock In {\em Proceedings of the 29th symposium on operating systems principles}, pages 611--626, 2023.

\bibitem{puzzle}
Kinman Lei, Yuyang Jin, Mingshu Zhai, Kezhao Huang, Haoxing Ye, and Jidong Zhai.
\newblock $\{$PUZZLE$\}$: Efficiently aligning large language models through $\{$Light-Weight$\}$ context switch.
\newblock In {\em 2024 USENIX Annual Technical Conference (USENIX ATC 24)}, pages 127--140, 2024.

\bibitem{TACO}
Rongao Li, Jie Fu, Bo-Wen Zhang, Tao Huang, Zhihong Sun, Chen Lyu, Guang Liu, Zhi Jin, and Ge~Li.
\newblock Taco: Topics in algorithmic code generation dataset.
\newblock {\em CoRR}, abs/2312.14852, 2023.

\bibitem{ddp}
Shen Li, Yanli Zhao, Rohan Varma, Omkar Salpekar, Pieter Noordhuis, Teng Li, Adam Paszke, Jeff Smith, Brian Vaughan, Pritam Damania, and Soumith Chintala.
\newblock Pytorch distributed: experiences on accelerating data parallel training.
\newblock {\em Proceedings of the VLDB Endowment}, 13:3005–3018, 2020.

\bibitem{li2021sequence}
Shenggui Li, Fuzhao Xue, Chaitanya Baranwal, Yongbin Li, and Yang You.
\newblock Sequence parallelism: Long sequence training from system perspective.
\newblock {\em arXiv preprint arXiv:2105.13120}, 2021.

\bibitem{math500}
Hunter Lightman, Vineet Kosaraju, Yura Burda, Harri Edwards, Bowen Baker, Teddy Lee, Jan Leike, John Schulman, Ilya Sutskever, and Karl Cobbe.
\newblock Let's verify step by step.
\newblock {\em arXiv preprint arXiv:2305.20050}, 2023.

\bibitem{what_if}
Jinkun Lin, Ziheng Jiang, Zuquan Song, Sida Zhao, Menghan Yu, Zhanghan Wang, Chenyuan Wang, Zuocheng Shi, Xiang Shi, Wei Jia, et~al.
\newblock Understanding stragglers in large model training using what-if analysis.
\newblock {\em arXiv preprint arXiv:2505.05713}, 2025.

\bibitem{deepseek}
Aixin Liu, Bei Feng, Bing Xue, Bingxuan Wang, Bochao Wu, Chengda Lu, Chenggang Zhao, Chengqi Deng, Chenyu Zhang, Chong Ruan, et~al.
\newblock Deepseek-v3 technical report.
\newblock {\em arXiv preprint arXiv:2412.19437}, 2024.

\bibitem{deepcoder}
Michael Luo, Sijun Tan, Roy Huang, Ameen Patel, Alpay Ariyak, Qingyang Wu, Xiaoxiang Shi, Rachel Xin, Colin Cai, Maurice Weber, Ce~Zhang, Li~Erran Li, Raluca~Ada Popa, and Ion Stoica.
\newblock Deepcoder: A fully open-source 14b coder at o3-mini level.
\newblock \url{https://pretty-radio-b75.notion.site/DeepCoder-A-Fully-Open-Source-14B-Coder-at-O3-mini-Level-1cf81902c14680b3bee5eb349a512a51}, 2025.
\newblock Notion Blog.

\bibitem{oreal}
Chengqi Lyu, Songyang Gao, Yuzhe Gu, Wenwei Zhang, Jianfei Gao, Kuikun Liu, Ziyi Wang, Shuaibin Li, Qian Zhao, Haian Huang, et~al.
\newblock Exploring the limit of outcome reward for learning mathematical reasoning.
\newblock {\em arXiv preprint arXiv:2502.06781}, 2025.

\bibitem{Synthetic1}
Justus Mattern, Sami Jaghouar, Manveer Basra, Jannik Straube, Matthew~Di Ferrante, Felix Gabriel, Jack~Min Ong, Vincent Weisser, and Johannes Hagemann.
\newblock Synthetic-1: Two million collaboratively generated reasoning traces from deepseek-r1, 2025.

\bibitem{mcp}
Model {C}ontext {P}rotocol --- official repository.
\newblock \url{https://github.com/modelcontextprotocol}.
\newblock Accessed: 2025-09-12.

\bibitem{realhf}
Zhiyu Mei, Wei Fu, Kaiwei Li, Guangju Wang, Huanchen Zhang, and Yi~Wu.
\newblock Realhf: Optimized rlhf training for large language models through parameter reallocation.
\newblock {\em arXiv e-prints}, pages arXiv--2406, 2024.

\bibitem{Ray}
Philipp Moritz, Robert Nishihara, Stephanie Wang, Alexey Tumanov, Richard Liaw, Eric Liang, Melih Elibol, Zongheng Yang, William Paul, Michael~I Jordan, et~al.
\newblock Ray: A distributed framework for emerging $\{$AI$\}$ applications.
\newblock In {\em 13th USENIX symposium on operating systems design and implementation (OSDI 18)}, pages 561--577, 2018.

\bibitem{PipeDream}
Deepak Narayanan, Aaron Harlap, Amar Phanishayee, Vivek Seshadri, Nikhil~R. Devanur, Gregory~R. Ganger, Phillip~B. Gibbons, and Matei Zaharia.
\newblock Pipedream: generalized pipeline parallelism for dnn training.
\newblock In {\em Proceedings of the 27th ACM Symposium on Operating Systems Principles}, SOSP '19. Association for Computing Machinery, 2019.

\bibitem{Efficient21}
Deepak Narayanan, Mohammad Shoeybi, Jared Casper, Patrick LeGresley, Mostofa Patwary, Vijay Korthikanti, Dmitri Vainbrand, Prethvi Kashinkunti, Julie Bernauer, Bryan Catanzaro, Amar Phanishayee, and Matei Zaharia.
\newblock Efficient large-scale language model training on gpu clusters using megatron-lm.
\newblock In {\em Proceedings of the International Conference for High Performance Computing, Networking, Storage and Analysis}, SC '21. Association for Computing Machinery, 2021.

\bibitem{nvidia2024cuda}
{NVIDIA Corporation}.
\newblock Cuda toolkit - free tools and training, 2024.
\newblock NVIDIA Developer. Accessed: 2024-09-19.

\bibitem{nccl}
{NVIDIA Corporation}.
\newblock Nvidia collective communications library (nccl), 2024.
\newblock Accessed: 2024-09-19.

\bibitem{o1}
OpenAI, :, Aaron Jaech, Adam Kalai, Adam Lerer, Adam Richardson, Ahmed El-Kishky, Aiden Low, Alec Helyar, Aleksander Madry, Alex Beutel, Alex Carney, Alex Iftimie, Alex Karpenko, Alex~Tachard Passos, Alexander Neitz, Alexander Prokofiev, Alexander Wei, Allison Tam, Ally Bennett, Ananya Kumar, Andre Saraiva, Andrea Vallone, Andrew Duberstein, Andrew Kondrich, Andrey Mishchenko, Andy Applebaum, Angela Jiang, Ashvin Nair, Barret Zoph, Behrooz Ghorbani, Ben Rossen, Benjamin Sokolowsky, Boaz Barak, Bob McGrew, Borys Minaiev, Botao Hao, Bowen Baker, Brandon Houghton, Brandon McKinzie, Brydon Eastman, Camillo Lugaresi, Cary Bassin, Cary Hudson, Chak~Ming Li, Charles de~Bourcy, Chelsea Voss, Chen Shen, Chong Zhang, Chris Koch, Chris Orsinger, Christopher Hesse, Claudia Fischer, Clive Chan, Dan Roberts, Daniel Kappler, Daniel Levy, Daniel Selsam, David Dohan, David Farhi, David Mely, David Robinson, Dimitris Tsipras, Doug Li, Dragos Oprica, Eben Freeman, Eddie Zhang, Edmund Wong, Elizabeth Proehl, Enoch Cheung, Eric
  Mitchell, Eric Wallace, Erik Ritter, Evan Mays, Fan Wang, Felipe~Petroski Such, Filippo Raso, Florencia Leoni, Foivos Tsimpourlas, Francis Song, Fred von Lohmann, Freddie Sulit, Geoff Salmon, Giambattista Parascandolo, Gildas Chabot, Grace Zhao, Greg Brockman, Guillaume Leclerc, Hadi Salman, Haiming Bao, Hao Sheng, Hart Andrin, Hessam Bagherinezhad, Hongyu Ren, Hunter Lightman, Hyung~Won Chung, Ian Kivlichan, Ian O'Connell, Ian Osband, Ignasi~Clavera Gilaberte, Ilge Akkaya, Ilya Kostrikov, Ilya Sutskever, Irina Kofman, Jakub Pachocki, James Lennon, Jason Wei, Jean Harb, Jerry Twore, Jiacheng Feng, Jiahui Yu, Jiayi Weng, Jie Tang, Jieqi Yu, Joaquin~Quiñonero Candela, Joe Palermo, Joel Parish, Johannes Heidecke, John Hallman, John Rizzo, Jonathan Gordon, Jonathan Uesato, Jonathan Ward, Joost Huizinga, Julie Wang, Kai Chen, Kai Xiao, Karan Singhal, Karina Nguyen, Karl Cobbe, Katy Shi, Kayla Wood, Kendra Rimbach, Keren Gu-Lemberg, Kevin Liu, Kevin Lu, Kevin Stone, Kevin Yu, Lama Ahmad, Lauren Yang, Leo Liu,
  Leon Maksin, Leyton Ho, Liam Fedus, Lilian Weng, Linden Li, Lindsay McCallum, Lindsey Held, Lorenz Kuhn, Lukas Kondraciuk, Lukasz Kaiser, Luke Metz, Madelaine Boyd, Maja Trebacz, Manas Joglekar, Mark Chen, Marko Tintor, Mason Meyer, Matt Jones, Matt Kaufer, Max Schwarzer, Meghan Shah, Mehmet Yatbaz, Melody~Y. Guan, Mengyuan Xu, Mengyuan Yan, Mia Glaese, Mianna Chen, Michael Lampe, Michael Malek, Michele Wang, Michelle Fradin, Mike McClay, Mikhail Pavlov, Miles Wang, Mingxuan Wang, Mira Murati, Mo~Bavarian, Mostafa Rohaninejad, Nat McAleese, Neil Chowdhury, Neil Chowdhury, Nick Ryder, Nikolas Tezak, Noam Brown, Ofir Nachum, Oleg Boiko, Oleg Murk, Olivia Watkins, Patrick Chao, Paul Ashbourne, Pavel Izmailov, Peter Zhokhov, Rachel Dias, Rahul Arora, Randall Lin, Rapha~Gontijo Lopes, Raz Gaon, Reah Miyara, Reimar Leike, Renny Hwang, Rhythm Garg, Robin Brown, Roshan James, Rui Shu, Ryan Cheu, Ryan Greene, Saachi Jain, Sam Altman, Sam Toizer, Sam Toyer, Samuel Miserendino, Sandhini Agarwal, Santiago Hernandez,
  Sasha Baker, Scott McKinney, Scottie Yan, Shengjia Zhao, Shengli Hu, Shibani Santurkar, Shraman~Ray Chaudhuri, Shuyuan Zhang, Siyuan Fu, Spencer Papay, Steph Lin, Suchir Balaji, Suvansh Sanjeev, Szymon Sidor, Tal Broda, Aidan Clark, Tao Wang, Taylor Gordon, Ted Sanders, Tejal Patwardhan, Thibault Sottiaux, Thomas Degry, Thomas Dimson, Tianhao Zheng, Timur Garipov, Tom Stasi, Trapit Bansal, Trevor Creech, Troy Peterson, Tyna Eloundou, Valerie Qi, Vineet Kosaraju, Vinnie Monaco, Vitchyr Pong, Vlad Fomenko, Weiyi Zheng, Wenda Zhou, Wes McCabe, Wojciech Zaremba, Yann Dubois, Yinghai Lu, Yining Chen, Young Cha, Yu~Bai, Yuchen He, Yuchen Zhang, Yunyun Wang, Zheng Shao, and Zhuohan Li.
\newblock Openai o1 system card.
\newblock {\em CoRR}, abs/2412.16720, 2024.

\bibitem{instructGPT}
Long Ouyang, Jeffrey Wu, Xu~Jiang, Diogo Almeida, Carroll Wainwright, Pamela Mishkin, Chong Zhang, Sandhini Agarwal, Katarina Slama, Alex Ray, et~al.
\newblock Training language models to follow instructions with human feedback.
\newblock {\em Advances in neural information processing systems}, 35:27730--27744, 2022.

\bibitem{verl-pipeline}
Agentica Project.
\newblock Verl-pipeline.
\newblock \url{https://github.com/agentica-project/verl-pipeline.git}, 2025.
\newblock Accessed: 2025-08-19.

\bibitem{torch-memory-snapshot}
{PyTorch Team}.
\newblock {\em PyTorch Memory Snapshot and Debugging CUDA Memory Usage}, 2023.
\newblock PyTorch documentation. Accessed: 2025-09-11.

\bibitem{Pollux}
Aurick Qiao, Sang~Keun Choe, Suhas~Jayaram Subramanya, Willie Neiswanger, Qirong Ho, Hao Zhang, Gregory~R. Ganger, and Eric~P. Xing.
\newblock Pollux: Co-adaptive cluster scheduling for goodput-optimized deep learning.
\newblock In {\em 15th {USENIX} Symposium on Operating Systems Design and Implementation}, OSDI '21, pages 1--18. {USENIX} Association, 2021.

\bibitem{mooncake}
Ruoyu Qin, Zheming Li, Weiran He, Mingxing Zhang, Yongwei Wu, Weimin Zheng, and Xinran Xu.
\newblock Mooncake: A kvcache-centric disaggregated architecture for llm serving.
\newblock {\em arXiv preprint arXiv:2407.00079}, 2024.

\bibitem{modserve}
Haoran Qiu, Anish Biswas, Zihan Zhao, Jayashree Mohan, Alind Khare, Esha Choukse, {\'I}{\~n}igo Goiri, Zeyu Zhang, Haiying Shen, Chetan Bansal, et~al.
\newblock Modserve: Scalable and resource-efficient large multimodal model serving.
\newblock {\em arXiv preprint arXiv:2502.00937}, 2025.

\bibitem{Pretrain}
Alec Radford, Karthik Narasimhan, Tim Salimans, and Ilya Sutskever.
\newblock Improving language understanding by generative pre-training.
\newblock Technical Report OpenAI TR-2018-01, OpenAI, San Francisco, CA, USA, 2018.

\bibitem{Zero}
Samyam Rajbhandari, Jeff Rasley, Olatunji Ruwase, and Yuxiong He.
\newblock Zero: Memory optimizations toward training trillion parameter models.
\newblock In {\em SC20: International Conference for High Performance Computing, Networking, Storage and Analysis}, pages 1--16. IEEE, 2020.

\bibitem{DeepSpeed}
Jeff Rasley, Samyam Rajbhandari, Olatunji Ruwase, and Yuxiong He.
\newblock Deepspeed: System optimizations enable training deep learning models with over 100 billion parameters.
\newblock In {\em Proceedings of the 26th ACM SIGKDD International Conference on Knowledge Discovery \& Data Mining}, KDD '20. Association for Computing Machinery, 2020.

\bibitem{Zero_offload}
Jie Ren, Samyam Rajbhandari, Reza~Yazdani Aminabadi, Olatunji Ruwase, Shuangyan Yang, Minjia Zhang, Dong Li, and Yuxiong He.
\newblock $\{$Zero-offload$\}$: Democratizing $\{$billion-scale$\}$ model training.
\newblock In {\em 2021 USENIX Annual Technical Conference (USENIX ATC 21)}, pages 551--564, 2021.

\bibitem{PPO}
John Schulman, Filip Wolski, Prafulla Dhariwal, Alec Radford, and Oleg Klimov.
\newblock Proximal policy optimization algorithms.
\newblock {\em arXiv preprint arXiv:1707.06347}, 2017.

\bibitem{grpo}
Zhihong Shao, Peiyi Wang, Qihao Zhu, Runxin Xu, Junxiao Song, Xiao Bi, Haowei Zhang, Mingchuan Zhang, YK~Li, Yang Wu, et~al.
\newblock Deepseekmath: Pushing the limits of mathematical reasoning in open language models.
\newblock {\em arXiv preprint arXiv:2402.03300}, 2024.

\bibitem{kvcache_preemption_1}
Haiying Shen, Tanmoy Sen, and Masahiro Tanaka.
\newblock Mitigating kv cache competition to enhance user experience in llm inference.
\newblock {\em arXiv preprint arXiv:2503.13773}, 2025.

\bibitem{verl}
Guangming Sheng, Chi Zhang, Zilingfeng Ye, Xibin Wu, Wang Zhang, Ru~Zhang, Yanghua Peng, Haibin Lin, and Chuan Wu.
\newblock Hybridflow: A flexible and efficient rlhf framework.
\newblock In {\em Proceedings of the Twentieth European Conference on Computer Systems}, pages 1279--1297, 2025.

\bibitem{megatron}
Mohammad Shoeybi, Mostofa Patwary, Raul Puri, Patrick LeGresley, Jared Casper, and Bryan Catanzaro.
\newblock Megatron-lm: Training multi-billion parameter language models using model parallelism.
\newblock {\em arXiv preprint arXiv:1909.08053}, 2019.

\bibitem{rl_survey}
Saksham~Sahai Srivastava and Vaneet Aggarwal.
\newblock A technical survey of reinforcement learning techniques for large language models.
\newblock {\em arXiv preprint arXiv:2507.04136}, 2025.

\bibitem{ZeroSearch}
Hao Sun, Zile Qiao, Jiayan Guo, Xuanbo Fan, Yingyan Hou, Yong Jiang, Pengjun Xie, Yan Zhang, Fei Huang, and Jingren Zhou.
\newblock Zerosearch: Incentivize the search capability of llms without searching.
\newblock {\em CoRR}, abs/2505.04588, 2025.

\bibitem{SFT1}
Rohan Taori, Ishaan Gulrajani, Tianyi Zhang, Yann Dubois, Xuechen Li, Carlos Guestrin, Percy Liang, and Tatsunori~B Hashimoto.
\newblock Stanford alpaca: An instruction-following llama model, 2023.

\bibitem{Gemini}
Gemini Team, Rohan Anil, Sebastian Borgeaud, Jean-Baptiste Alayrac, Jiahui Yu, Radu Soricut, Johan Schalkwyk, Andrew~M Dai, Anja Hauth, Katie Millican, et~al.
\newblock Gemini: a family of highly capable multimodal models.
\newblock {\em arXiv preprint arXiv:2312.11805}, 2023.

\bibitem{kimi1_5}
Kimi Team, Angang Du, Bofei Gao, Bowei Xing, Changjiu Jiang, Cheng Chen, Cheng Li, Chenjun Xiao, Chenzhuang Du, Chonghua Liao, et~al.
\newblock Kimi k1. 5: Scaling reinforcement learning with llms.
\newblock {\em arXiv preprint arXiv:2501.12599}, 2025.

\bibitem{kimi_vl}
Kimi Team, Angang Du, Bohong Yin, Bowei Xing, Bowen Qu, Bow~en Wang, Cheng Chen, Chenlin Zhang, Chenzhuang Du, Chu Wei, et~al.
\newblock Kimi-vl technical report.
\newblock {\em arXiv preprint arXiv:2504.07491}, 2025.

\bibitem{qwen2.5}
Qwen Team.
\newblock Qwen2.5: A party of foundation models, September 2024.

\bibitem{slime2024}
THUDM.
\newblock slime: A llm post-training framework for rl scaling, 2025.
\newblock GitHub repository.

\bibitem{Attention17}
Ashish Vaswani, Noam Shazeer, Niki Parmar, Jakob Uszkoreit, Llion Jones, Aidan~N. Gomez, Łukasz Kaiser, and Illia Polosukhin.
\newblock Attention is all you need.
\newblock In {\em Advances in Neural Information Processing Systems}, NeurIPS '17, 2017.

\bibitem{trl}
Leandro von Werra, Younes Belkada, Lewis Tunstall, Edward Beeching, Tristan Thrush, Nathan Lambert, Shengyi Huang, Kashif Rasul, and Quentin Gallouédec.
\newblock Trl: Transformer reinforcement learning.
\newblock \url{https://github.com/huggingface/trl}, 2020.

\bibitem{Tenplex}
Marcel Wagenländer, Guo Li, Bo~Zhao, Luo Mai, and Peter Pietzuch.
\newblock Tenplex: Dynamic parallelism for deep learning using parallelizable tensor collections.
\newblock In {\em Proceedings of the ACM SIGOPS 30th Symposium on Operating Systems Principles}, SOSP '24. Association for Computing Machinery, 2024.

\bibitem{kvcacheinthewild}
Jiahao Wang, Jinbo Han, Xingda Wei, Sijie Shen, Dingyan Zhang, Chenguang Fang, Rong Chen, Wenyuan Yu, and Haibo Chen.
\newblock Kvcache cache in the wild: Characterizing and optimizing kvcache cache at a large cloud provider.
\newblock {\em arXiv preprint arXiv:2506.02634}, 2025.

\bibitem{burstgpt}
Yuxin Wang, Yuhan Chen, Zeyu Li, Xueze Kang, Yuchu Fang, Yeju Zhou, Yang Zheng, Zhenheng Tang, Xin He, Rui Guo, et~al.
\newblock Burstgpt: A real-world workload dataset to optimize llm serving systems.
\newblock In {\em Proceedings of the 31st ACM SIGKDD Conference on Knowledge Discovery and Data Mining V. 2}, pages 5831--5841, 2025.

\bibitem{mlaas}
Qizhen Weng, Wencong Xiao, Yinghao Yu, Wei Wang, Cheng Wang, Jian He, Yong Li, Liping Zhang, Wei Lin, and Yu~Ding.
\newblock $\{$MLaaS$\}$ in the wild: Workload analysis and scheduling in $\{$Large-Scale$\}$ heterogeneous $\{$GPU$\}$ clusters.
\newblock In {\em 19th USENIX Symposium on Networked Systems Design and Implementation (NSDI 22)}, pages 945--960, 2022.

\bibitem{servegen}
Yuxing Xiang, Xue Li, Kun Qian, Wenyuan Yu, Ennan Zhai, and Xin Jin.
\newblock Servegen: Workload characterization and generation of large language model serving in production.
\newblock {\em arXiv preprint arXiv:2505.09999}, 2025.

\bibitem{flexrlhf}
Youshao Xiao, Zhenglei Zhou, Fagui Mao, Weichang Wu, Shangchun Zhao, Lin Ju, Lei Liang, Xiaolu Zhang, and Jun Zhou.
\newblock Flexrlhf: A flexible placement and parallelism framework for efficient rlhf training.
\newblock In {\em 2025 IEEE International Parallel and Distributed Processing Symposium (IPDPS)}, pages 358--369. IEEE, 2025.

\bibitem{xiaomi2025mimo}
LLM Xiaomi, Bingquan Xia, Bowen Shen, Dawei Zhu, Di~Zhang, Gang Wang, Hailin Zhang, Huaqiu Liu, Jiebao Xiao, Jinhao Dong, et~al.
\newblock Mimo: Unlocking the reasoning potential of language model--from pretraining to posttraining.
\newblock {\em arXiv preprint arXiv:2505.07608}, 2025.

\bibitem{HotpotQA}
Zhilin Yang, Peng Qi, Saizheng Zhang, Yoshua Bengio, William~W. Cohen, Ruslan Salakhutdinov, and Christopher~D. Manning.
\newblock Hotpotqa: A dataset for diverse, explainable multi-hop question answering.
\newblock {\em CoRR}, abs/1809.09600, 2018.

\bibitem{kvcache_preemption_2}
Jie Ye, Jaime Cernuda, Avinash Maurya, Xian-He Sun, Anthony Kougkas, and Bogdan Nicolae.
\newblock Characterizing the behavior and impact of kv caching on transformer inferences under concurrency.
\newblock In {\em 2025 IEEE International Parallel and Distributed Processing Symposium (IPDPS)}, pages 1191--1202. IEEE, 2025.

\bibitem{dapo}
Qiying Yu, Zheng Zhang, Ruofei Zhu, Yufeng Yuan, Xiaochen Zuo, Yu~Yue, Weinan Dai, Tiantian Fan, Gaohong Liu, Lingjun Liu, et~al.
\newblock Dapo: An open-source llm reinforcement learning system at scale.
\newblock {\em arXiv preprint arXiv:2503.14476}, 2025.

\bibitem{turbomind}
Li~Zhang, Youhe Jiang, Guoliang He, Xin Chen, Han Lv, Qian Yao, Fangcheng Fu, and Kai Chen.
\newblock Efficient mixed-precision large language model inference with turbomind.
\newblock {\em CoRR}, abs/2508.15601, 2025.

\bibitem{fsdp}
Yanli Zhao, Andrew Gu, Rohan Varma, Liang Luo, Chien-Chin Huang, Min Xu, Less Wright, Hamid Shojanazeri, Myle Ott, Sam Shleifer, et~al.
\newblock Pytorch fsdp: experiences on scaling fully sharded data parallel.
\newblock {\em arXiv preprint arXiv:2304.11277}, 2023.

\bibitem{BlendServe}
Yilong Zhao, Shuo Yang, Kan Zhu, Lianmin Zheng, Baris Kasikci, Yang Zhou, Jiarong Xing, and Ion Stoica.
\newblock Blendserve: Optimizing offline inference for auto-regressive large models with resource-aware batching.
\newblock {\em CoRR}, abs/2411.16102, 2024.

\bibitem{Alpa}
Lianmin Zheng, Zhuohan Li, Hao Zhang, Yonghao Zhuang, Zhifeng Chen, Yanping Huang, Yida Wang, Yuanzhong Xu, Danyang Zhuo, Eric~P. Xing, Joseph~E. Gonzalez, and Ion Stoica.
\newblock Alpa: Automating inter- and {Intra-Operator} parallelism for distributed deep learning.
\newblock In {\em 16th USENIX Symposium on Operating Systems Design and Implementation}, OSDI '22, pages 559--578. USENIX Association, 2022.

\bibitem{sglang}
Lianmin Zheng, Liangsheng Yin, Zhiqiang Xie, Chuyue~Livia Sun, Jeff Huang, Cody~Hao Yu, Shiyi Cao, Christos Kozyrakis, Ion Stoica, Joseph~E Gonzalez, et~al.
\newblock Sglang: Efficient execution of structured language model programs.
\newblock {\em Advances in neural information processing systems}, 37:62557--62583, 2024.

\bibitem{BatchLLM}
Zhen Zheng, Xin Ji, Taosong Fang, Fanghao Zhou, Chuanjie Liu, and Gang Peng.
\newblock Batchllm: Optimizing large batched llm inference with global prefix sharing and throughput-oriented token batching.
\newblock {\em CoRR}, abs/2412.03594, 2025.

\bibitem{deepeyes}
Ziwei Zheng, Michael Yang, Jack Hong, Chenxiao Zhao, Guohai Xu, Le~Yang, Chao Shen, and Xing Yu.
\newblock Deepeyes: Incentivizing "thinking with images" via reinforcement learning, 2025.

\bibitem{DistServe}
Yinmin Zhong, Shengyu Liu, Junda Chen, Jianbo Hu, Yibo Zhu, Xuanzhe Liu, Xin Jin, and Hao Zhang.
\newblock {DistServe}: Disaggregating prefill and decoding for goodput-optimized large language model serving.
\newblock In {\em 18th USENIX Symposium on Operating Systems Design and Implementation}, OSDI '24, pages 193--210. USENIX Association, 2024.

\bibitem{streamrl}
Yinmin Zhong, Zili Zhang, Xiaoniu Song, Hanpeng Hu, Chao Jin, Bingyang Wu, Nuo Chen, Yukun Chen, Yu~Zhou, Changyi Wan, et~al.
\newblock Streamrl: Scalable, heterogeneous, and elastic rl for llms with disaggregated stream generation.
\newblock {\em arXiv preprint arXiv:2504.15930}, 2025.

\bibitem{rlhfuse}
Yinmin Zhong, Zili Zhang, Bingyang Wu, Shengyu Liu, Yukun Chen, Changyi Wan, Hanpeng Hu, Lei Xia, Ranchen Ming, Yibo Zhu, et~al.
\newblock Optimizing rlhf training for large language models with stage fusion.
\newblock {\em arXiv preprint arXiv:2409.13221}, 2024.

\end{thebibliography}

\newpage

\appendix


\begin{figure*}[t]
\begin{center}
\normalsize
\begin{verbatim}
verl.protocol.DataProto
    --batch(TensorDict)
    ----input_ids:Tensor(int64, shape=(bsz, max_prompt_len))
    ----attention_mask: torch.Size([64, 16000]) (torch.int64)
    ----position_ids: torch.Size([64, 16000]) (torch.int64)
    
    --non_tensor_batch
    ----data_source(str)
    ----ability(str)
    ----reward_model(dict{'ground_truth': 'A', 'style': 'rule'}) % rule-based reward
    ----extra_info(dict, keys=('answer', 'index', 'question', 'split')) % extra information
    ----multi_modal_data(dict) % multi-modal data
    ------key: ['video'] 4-D array % key: video 4-dimensional array
    ------video: <class 'list'> length=1, dtype=float32) % video: list type, length=1, dtype=float32
    ----multi_modal_inputs(dict) % multi-modal inputs
    ------key: ['pixel_values_videos', 'video_grid_thw']
    ------value:{pixel_values_videos: <class 'torch.Tensor'> shape=torch.Size([3840, 1176], float32)
            video_grid_thw: <class 'torch.Tensor'> shape=torch.Size([1, 3], int64)[[ 9, 36, 46]]}
    ----raw_prompt_ids: 1-D int list % raw prompt IDs: 1-dimensional integer list
    ----index: int
    ----tools_kwargs: None
\end{verbatim}
\end{center}
\caption{An example of the data protocol in RL training system.}
\label{fig:trace_record}
\end{figure*}

\section{Representative RL tasks}\label{repre_algo}
\begin{table}[t]
\centering
\renewcommand{\arraystretch}{1.4}
\resizebox{\linewidth}{!}
{
\begin{tabular}{l l l l}
\toprule
\textbf{Task} & \textbf{Model size} & \textbf{Total Step} & \textbf{Dataset} \\
\midrule
Mathematics & 32B\cite{qwen2.5} & 191 & DAPO\cite{dapo} \\
programming & 14B\cite{deepseekai2025deepseekr1incentivizingreasoningcapability} & 190 & DeepCoder\cite{deepcoder} \\
Searching & 7B\cite{qwen2.5} & 147 & NQ\_hotpotqa\cite{search_r1} \\
Video Understanding & 7B\cite{video_r1} & 112 & RoboVQA\cite{cosmos} \\
Image Understanding & 235B & 46 & In house \\
Mathematics & 235B & 188 & In house \\
Tool Use & 235B & 59 & In house \\
\bottomrule
\end{tabular}
}
\caption{Summary of \TN.}
\label{tab:rl-traces}
\end{table}

In recent months, RL algorithm researches have emerged, spanning mathematics, programming, tool use, and multimodal content. We select several representative algorithms to conduct detailed system analysis in Section \ref{sec_profile}.

\emph{Mathematics.} To enhance LLM reasoning capabilities, industry practice commonly applies reinforcement learning on math-related datasets. DAPO\cite{dapo} builds on the GRPO\cite{grpo} algorithm with dynamic sampling policy and provides a high-quality mathematics dataset. We adopt DAPO as the representative algorithm for mathematical tasks.

\emph{Programming.} A common approach to endow LLMs with strong programming and reasoning ability is to perform RL using coding datasets. DeepCoder\cite{deepcoder} curates high-quality open-source data and applies rigorous filtering. After filtering, the dataset contains 24K high-quality coding problems: 7.5K from TACO Verified\cite{TACO}, 16K from PrimeIntellect's SYNTHETIC-1\cite{Synthetic1}, and 600 from LiveCodeBench\cite{LiveCodeBench}. We use this algorithm to represent programming tasks.

\emph{Searching.} Using external tools enables LLMs to generate more reasonable and reliable responses. To equip LLMs with tool using capacity, many methods have been proposed to improve LLM competence with external tools—such as web search for answering questions, program execution\cite{retool} for producing verifiable outputs, and image-processing utilities. Search R1\cite{search_r1} is a highly influential project in searching RL tasks which merges the NQ\cite{nq} and HotpotQA\cite{HotpotQA} training sets to form a unified dataset for RL training. We select it as the representative one.

\emph{Video understanding.} To enable LLMs to understand video information, numerous algorithms and datasets\cite{rl_survey} aim to strengthen LLM performance across multimodal tasks. In particular, the RoboVQA dataset\cite{cosmos} trains LLMs to comprehend video content and issue instructions for a robot's next actions. We treat it as the representative for video understanding.

\section{Related Work}
\textbf{RL training systems.} Spurred by rapid algorithmic progress, many RL training systems have appeared in recent months \cite{openrlhf,trl}. Verl\cite{verl} proposes a single-controller/multi-controller architecture to ensure both usability and efficiency. RealHF\cite{realhf} leverages careful resource allocation to avoid expensive communication costs. Some systems \cite{puzzle, rlhfuse} conduct inter- and intra-stage optimizations to shrink GPU idle time. Furthermore, previous works\cite{flexrlhf,streamrl,asyncflow,verl-pipeline} model multi-stage RL training as a pipeline for acceleration. Beyond this, algorithm–system co-design, such as partial rollout\cite{slime2024,kimi1_5}, asynchronous RL training\cite{areal, xiaomi2025mimo,streamrl} has been explored to mitigate the effects of long-tailed sample distributions. Other acceleration techniques, including load-aware scheduling\cite{streamrl} and elastic scaling\cite{rlhfuse}, have also been explored.

\textbf{Trace analysis.}
Prior work has analyzed DL workloads across different organizations \cite{Philly,helios}. Acme\cite{acme} from Shanghai AI Lab presents a multifaceted study of cluster jobs predominantly running LLM workloads, covering utilization and fault tolerance. ServeGen\cite{servegen} provides a detailed analysis of request patterns in Alibaba’s internal LLM services—including text, multimodal, and reasoning workloads—and designs a system for data desensitization. KV Cache in the Wild\cite{kvcacheinthewild} analyzes KV-cache traces from Alibaba’s LLM services and proposes a more effective management system. Alibaba’s PAI\cite{mlaas} contributes to this discourse by analyzing the challenges encountered within their clusters from both temporal and spatial viewpoints. Concurrently, MegaScale\cite{MegaScale} and ByteScale\cite{bytescale} present ByteDance’s experience in training LLMs using a formidable array of over 10,000.

\end{document}